\documentclass{gammarobot}
\usepackage[utf8]{inputenc}
\usepackage{amsmath,amssymb,amsfonts,amsthm,mathtools}
\usepackage{tabularx,makecell,adjustbox}
\usepackage[ruled,vlined]{algorithm2e}
\usepackage{fancyhdr}
\fancypagestyle{gammaTeleAI}{%
  \fancyhf{}
  \fancyhead[C]{\small\sffamily Gamma Robotics / TeleAI}
  \fancyfoot[C]{\thepage}
  \renewcommand{\headrulewidth}{\ifnum\value{page}>1 0.3pt\else 0pt\fi}
  
}
\makeatletter
\let\ps@plain\ps@gammaTeleAI
\makeatother
\DeclareRobustCommand{\projecticon}{%
  \makebox[1.35em][c]{\raisebox{-0.18em}{\includegraphics[height=1.05em]{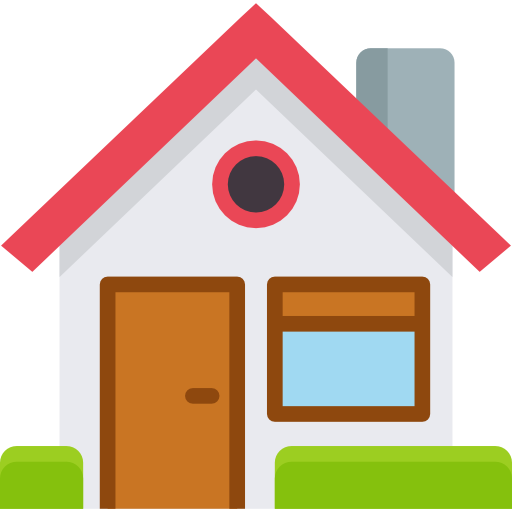}}}\hspace{0.35em}%
}
\renewcommand\project[1]{\metadata[\projecticon Project]{#1}}

\let\cite\citep
\newcommand{\system}{\textsc{LocoManiLab}}
\newcommand{\IEEEPARstart}[2]{#1#2}
\BeforeBeginEnvironment{tabular}{\begin{adjustbox}{max width=\linewidth}}
\AfterEndEnvironment{tabular}{\end{adjustbox}}
\let\reportoriginalincludegraphics\includegraphics
\NewDocumentCommand{\reportcolumnimage}{s O{} m}{%
  \adjustbox{max width=0.70\linewidth}{%
    \IfBooleanTF{#1}{\reportoriginalincludegraphics*[#2]{#3}}{\reportoriginalincludegraphics[#2]{#3}}%
  }%
}
\AtBeginEnvironment{figure}{\let\includegraphics\reportcolumnimage}

\pretocmd{\bibliographystyle}{\FloatBarrier}{}{}
\newcommand{\appendices}{%
  \clearpage
  \beginappendix
  \suppressfloats[t]
  \pretocmd{\section}{\FloatBarrier}{}{}
  \renewcommand{\thesubsection}{\thesection-\Alph{subsection}}%
  \renewcommand{\thesubsubsection}{\thesubsection\arabic{subsubsection}}%
}
\hypersetup{
  pdfauthor={Dewei Wang, Jinrui Han, Jiyuan Shi, Chenyun Zhang, Xianbo Lu, Jiabao Wu, Jiqin Feng, Jipeng Kong, Xinzhe Liu, Ping Luo, Yuchen Xiao, Xingxing Wang, Chenjia Bai, Xuelong Li},
  pdftitle={LocoManiLab: A Unified Testbed for Data, Policies, Controllers, and Deployment in Humanoid Loco-Manipulation},
  pdfkeywords={Humanoid Robots, Imitation Learning, Performance Evaluation and Benchmarking, Humanoid Loco-Manipulation}
}

\renewcommand\authorformat[2][]{\mbox{\small{\sffamily\bfseries #2}$^{#1}$}}
\renewcommand\affiliation[2][]{\addtolist[#1]{#2}{\affiliationlist}{\affiliationformat}{\par}}
\renewcommand\affiliationformat[2][]{{\small $^{#1}$#2}}

\title{\system: A Unified Testbed for Data, Policies, Controllers, and Deployment in Humanoid Loco-Manipulation}
\author[1,2,*,\S]{Dewei Wang}
\author[1,3,*,\S]{Jinrui Han}
\author[1,*]{Jiyuan Shi}
\author[1]{Chenyun Zhang}
\author[1,4]{Xianbo Lu}
\author[5]{Jiabao Wu}
\author[1,2]{Jiqin Feng}
\author[6]{Jipeng Kong}
\author[1]{Xinzhe Liu}
\author[3]{Ping Luo}
\author[7]{Yuchen Xiao}
\author[7]{Xingxing Wang}
\author[1,8,\dagger]{Chenjia Bai}
\author[1,\dagger]{Xuelong Li}
\affiliation[1]{Institute of Artificial Intelligence (TeleAI), China Telecom}
\affiliation[2]{University of Science and Technology of China}
\affiliation[3]{The University of Hong Kong}
\affiliation[4]{Zhejiang University}
\affiliation[5]{Shanghai Jiao Tong University}
\affiliation[6]{ShanghaiTech University}
\affiliation[7]{Unitree Robotics}
\affiliation[8]{Gamma Robotics}
\contribution{\textsuperscript{*}Equal contribution\quad\textsuperscript{\S}Co-project leaders\quad\textsuperscript{$\dagger$}Corresponding authors}

\date{\today}
\project{\url{https://locomanilab.github.io/}}

\abstract{
Humanoid robots must coordinate locomotion and manipulation in everyday environments, yet progress is difficult to assess across differing policies, controllers, and task settings.
\system{} provides an integrated testbed spanning data collection and generation, policy training, evaluation, and physical deployment.
Its task suite covers 234 atomic and compositional tasks across five household scene categories, supported by 51.5 hours of simulated teleoperation.
The evaluation covers twelve policies with a shared controller across task and scene splits and perturbation levels, two policies across five controllers, and studies of demonstration scale and data-source mixtures.
Success rates fall on unseen tasks and scenes and generally decline as visual perturbations increase, with differences in generalization and robustness across policies. For data generation, a humanoid motion generation model is fine-tuned for loco-manipulation to study the alignment of its outputs with task execution requirements and its integration into the data generation pipeline. Three simulated loco-manipulation tasks closely match real-world setups, enabling evaluation of three policies in both domains. Similar policy rankings support the reliability of simulation-based evaluation.
}

\begin{document}
\maketitle


\begin{figure*}[!t]
  \centering
  \includegraphics[width=\textwidth]{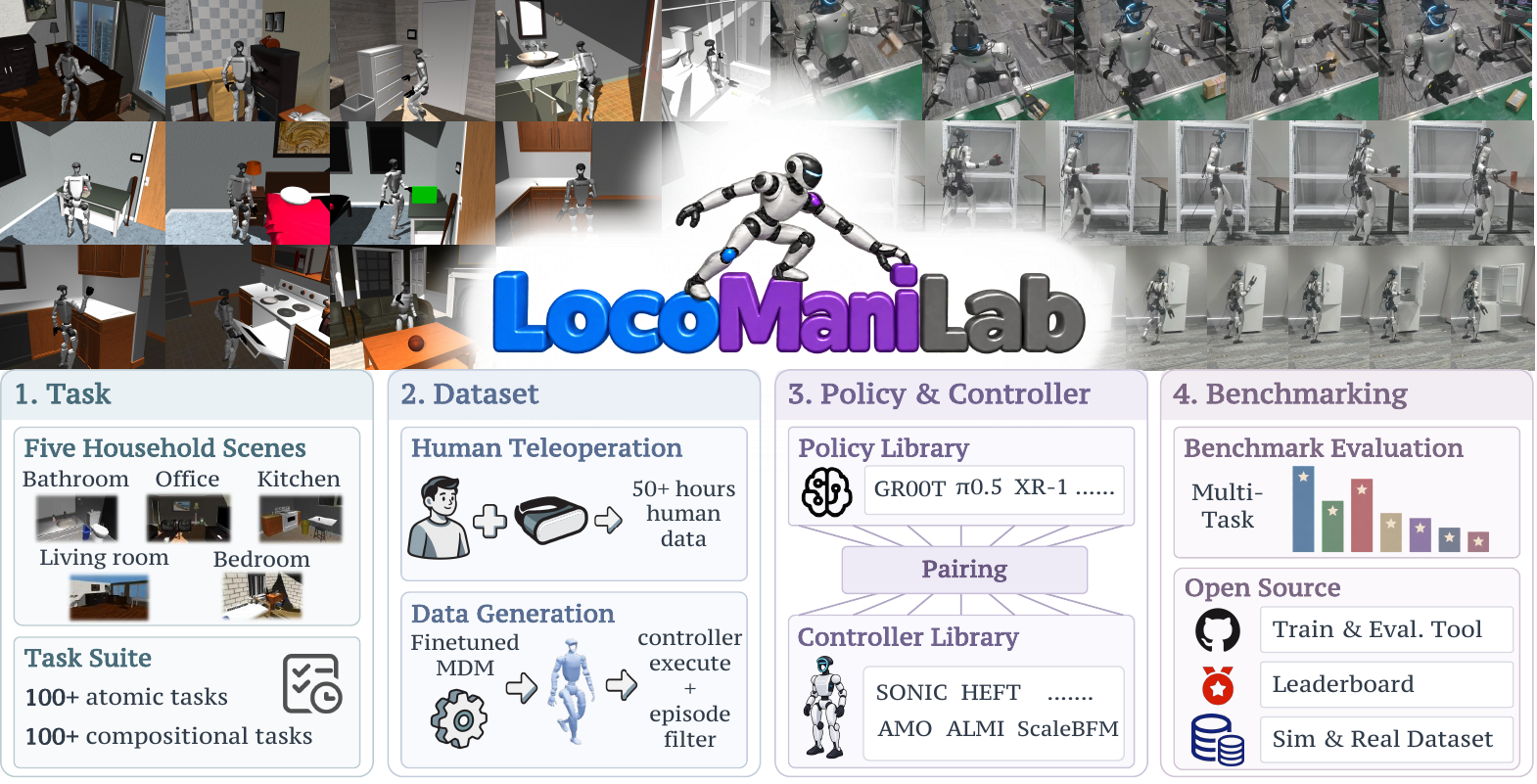}
  \caption{Overview of \system{}. Scene categories span kitchens, living rooms, bedrooms, bathrooms, and offices, with affordances including articulated appliances, drawers, object placement, and desk interactions. Data collection combines teleoperation with controller-executed motion generation, while dedicated interfaces connect policies to compatible controllers. Evaluation measures task completion and execution quality on seen tasks, unseen tasks and scenes, and under domain randomization, with a shared workflow.}
  \label{fig:overview}
\end{figure*}

\section{Introduction}\label{sec:introduction}

\IEEEPARstart{H}{umanoid} robots combine navigation, posture changes, and manipulation to act throughout human environments.
Most systems use a high-level policy to map vision and language to intermediate commands~\cite{wei2026psi0,shi2026egohumanoid,hu2026openhlm,jiang2025wholebodyvla}, which a whole-body controller realizes under balance, contact, and actuator constraints~\cite{luo2025sonic,li2025amo,shi2025almi,liu2026heft}.
Evaluated behavior consequently depends on the policy--controller pairing, its command interface, and the data used to train it.
Evaluating humanoid loco-manipulation across different policy--controller pairings requires a large, diverse set of humanoid-specific tasks and a comprehensive evaluation platform with standardized protocols.

Existing work has advanced humanoid evaluation through broader task coverage~\cite{sferrazza2024humanoidbench,wei2026simple}, demonstration generation~\cite{mandlekar2026humanoidmimicgen}, physical deployment~\cite{hu2026openhlm}, and transfer across controllers~\cite{wang2026humanoidarena}. However, these complementary efforts leave a need for a unified framework that assesses performance across a large and diverse set of tasks and execution quality while accounting for policy, controller, and demonstration choices. Similar task-success rates can conceal differences in stability and effort, while performance on familiar tasks alone does not establish generalization to new tasks or scenes. 
Household tasks provide a representative setting for such evaluation because they closely couple navigation, posture adaptation, and manipulation. Opening a drawer, for example, requires the robot to approach the handle, establish a suitable stance, and maintain stable contact during manipulation. Long-horizon tasks further require skills to compose through the physical states left by preceding actions. Successful navigation can still leave the robot poorly positioned for the subsequent reach or manipulation. These demands make household tasks well suited to evaluating both individual skills and their composition, together with the motion quality arising from policy--controller coordination. Motivated by these requirements, \system{} provides a comprehensive humanoid loco-manipulation testbed that combines diverse household tasks with standardized evaluation protocols to study policies, their execution interfaces between controllers, and their dependence on demonstrations. Fig.~\ref{fig:overview} summarizes the testbed and its four main contributions, outlined below.

\textbf{Unified testbed.}
The testbed provides a large and diverse suite of humanoid loco-manipulation tasks. A shared episode format and controller-specific adapters integrate data collection and generation, policy training, controller execution, and evaluation across simulation and hardware.

\textbf{Policy and controller.}
Twelve policies, including vision-language-action (VLA) models, world-action models (WAMs), and an imitation-learning baseline, are evaluated under a shared whole-body controller.
A complementary study evaluates two VLA models across five whole-body controllers, assessing both task completion and execution quality.

\textbf{Tasks and demonstrations.}
The task suite comprises 234 atomic and compositional tasks across five household scene categories, supported by 51.5 hours of simulated teleoperation.
Motion synthesis and trajectory augmentation provide additional demonstration sources, whose downstream utility is evaluated separately from data generator adaptation.

\textbf{Benchmarking and deployment.}
Evaluation distinguishes familiar tasks, unseen task definitions, unseen scenes, and visual randomization, with dedicated studies of demonstration scale and source mixtures.
Three policies are evaluated on three matched tasks in simulation and the real world.

The results show that policy rankings on familiar tasks largely hold on unseen tasks and scenes, whereas absolute success drops sharply under these shifts.
Policy rankings can change with perturbation level and controller choice, while higher task success does not necessarily imply better motion quality.
The benefits of additional demonstrations and mixed data sources vary across policies and tasks, while fine-tuning the motion generator improves motion generation and task success.
These findings motivate evaluating generalization, execution quality, and data utility as distinct but connected dimensions.
Table~\ref{tab:related_comparison} compares \system{} with existing simulation-based humanoid loco-manipulation evaluation platforms across key dimensions, including demonstration scale, task suite, policy and controller coverage, data generation, and hardware deployment.

\begin{table*}[t]
  \caption{Comparison of humanoid loco-manipulation systems.}
  \label{tab:related_comparison}
  \centering
  \footnotesize
  \setlength{\tabcolsep}{5pt}
  \begin{tabular}{@{}lccccccc@{}}
    \toprule
    System & \begin{tabular}[c]{@{}c@{}}Sim data (h)\end{tabular} &
    \begin{tabular}[c]{@{}c@{}}Real data (h)\end{tabular} &
Sim tasks & \begin{tabular}[c]{@{}c@{}}Policies\end{tabular} &
    \begin{tabular}[c]{@{}c@{}}Controllers\end{tabular} & Data-gen. &
    \begin{tabular}[c]{@{}c@{}}Hardware deployment\end{tabular} \\
    \midrule
    HumanoidBench~\cite{sferrazza2024humanoidbench} & -- & -- & 27 & -- & -- & -- & -- \\
    SIMPLE~\cite{wei2026simple} & 3.0 & 7.9 & 60 & 9 & 2 & \checkmark & \checkmark \\
    HumanoidArena~\cite{wang2026humanoidarena} & $\sim$5.0 & -- & 7 & 4 & 2 & -- & -- \\
    OpenHLM~\cite{hu2026openhlm} & -- & -- & -- & 3 & 2 & -- & \checkmark \\
    \system{} & 51.5 & 9.0 & 234 & 12 & 5 & \checkmark & \checkmark \\
    \bottomrule
  \end{tabular}
\end{table*}

\section{Related Work}
\label{sec:related_work}

\subsection{Humanoid Whole-Body Controllers}
Whole-body controllers enable pose-based teleoperation~\cite{he2024omnih2o,he2024hover}, decoupled locomotion and manipulation~\cite{li2025amo,shi2025almi}, and general motion tracking through diverse training references and architectures~\cite{chen2025gmt,han2025kungfubot2,luo2025sonic,liu2026heft,zeng2026scalebfm}.
Recent work also distills specialist skills into a vision-based generalist locomotion policy~\cite{wang2026xloco}.
Related studies investigate whole-body bilateral teleoperation and learning task priorities from demonstrations for humanoid loco-manipulation~\cite{ramos2018humanoidteleop,silverio2019taskpriorities}, as well as object-aware mobile manipulation with a wheeled humanoid~\cite{baek2026wholebodyteleop}.
Across these systems, command interfaces range from velocity and arm targets to full-body references and learned motion tokens, shaping both policy training and execution.
These interfaces determine how motion decisions are divided between the policy and controller. Velocity commands delegate gait control to the controller, whereas control through reference motions requires denser and richer control signals.
Comparing these systems therefore requires reporting command representations alongside task success and motion quality.
\system{} makes these interfaces explicit and supports policy evaluation with five whole-body controllers.

\subsection{Humanoid Whole-Body Loco-Manipulation}
Motion-driven methods acquire humanoid loco-manipulation skills from human motions and interaction demonstrations~\cite{weng2025hdmi,wang2025physhsi,lin2026prohoi}.
Physics-aware whole-body control also supports dynamic humanoid--object interaction such as skateboarding~\cite{han2026husky}.
Generation-based approaches predict whole-body motion for execution through tracking controllers~\cite{wang2026vlk}, while planning-based methods construct loco-manipulation trajectories for policy learning~\cite{mandlekar2026humanoidmimicgen}.
General-purpose VLAs transfer visual and language representations to robot action~\cite{physicalintelligence2025pi05,bjorck2025groot,shukor2025smolvla,zheng2025xvla,chen2025internvla}.
Humanoid VLA systems extend this approach to whole-body loco-manipulation, including learning from egocentric human video~\cite{wei2026psi0,shi2026egohumanoid,jiang2025wholebodyvla}. WAMs offer another route to general-purpose control by jointly learning visual dynamics and actions~\cite{ye2026dreamzero,yuan2026fastwam}.
Despite these advances, comprehensive evaluation of VLA and WAM-based humanoid loco-manipulation systems remains limited.
\system{} systematically evaluates policies under shared task protocols, assessing task success, generalization, and execution quality across controller and training data configurations.

\subsection{Benchmarks for Manipulation and Loco-Manipulation}
Manipulation benchmarks provide reusable task suites, diverse scenes, and protocols for evaluating generalization~\cite{james2019rlbench,liu2023libero,nasiriany2024robocasa,tao2024maniskill3,nasiriany2026robocasa365}.
Cross-embodiment datasets and simulation--hardware comparisons highlight the importance of explicit data and evaluation conventions~\cite{openx2023,li2024simpler}, while benchmark audits show that model rankings can depend on action representation and environment~\cite{guruprasad2024vla}.
Complementary studies develop simulation-to-real evaluation protocols for contact-rich visuotactile manipulation and use imperfect simulation to reduce the physical trials required for policy tuning~\cite{chen2024tactilesim2real,he2025simulationaided}.
For humanoid robots, HumanoidBench~\cite{sferrazza2024humanoidbench} covers locomotion and whole-body manipulation, SIMPLE~\cite{wei2026simple} evaluates multiple policy families, including imitation-learning baselines, and HumanoidArena~\cite{wang2026humanoidarena} examines controller-dependent transfer.
OpenHLM~\cite{hu2026openhlm} further evaluates policies and controllers on hardware.
Together, these studies highlight the need to report both task coverage and the execution configurations underlying benchmark scores.
\system{} brings policy coverage, controller-dependent execution, and demonstration choices into a shared evaluation framework.

\section{\system{}}
\label{sec:method}

\subsection{Overview}

\system{} provides a unified testbed for humanoid loco-manipulation, connecting demonstration acquisition, policy learning, controller execution, and evaluation. The task suite covers both atomic and compositional behaviors, supported by teleoperated demonstrations, motion synthesis, and trajectory augmentation.
Shared data workflows and policy--controller adapters connect these demonstration sources to training and deployment while accommodating different observation and action representations. 
The benchmark compares twelve policies spanning VLAs, WAMs, and an imitation-learning baseline under a shared whole-body controller, complemented by a study of two VLA policies across five controllers.
Evaluation measures both task completion and motion quality, examining generalization to unseen tasks and scenes, robustness to visual perturbations, and the influence of controller choice on execution. 
Further studies investigate demonstration scale, data-source mixtures, and motion generator fine-tuning, while corresponding simulation and real-world tasks provide evidence of policy performance across both domains. 
Together, these components support systematic comparisons of humanoid loco-manipulation systems and the data and execution choices underlying their performance.
To support these evaluations, the benchmark includes simulated household environments with diverse room layouts and interactive objects. The scenes run in MuJoCo~\cite{todorov2012mujoco} and combine objects and fixtures from RoboCasa~\cite{nasiriany2024robocasa} and Objaverse~\cite{deitke2022objaverse} with room layouts from MolmoSpaces' iTHOR-derived scenes~\cite{kim2026molmospaces} and Holodeck offices~\cite{yang2024holodeck}.

\subsection{Policies and Controllers}
\label{par:low_level}

\system{} evaluates twelve policies across three categories, comprising ten pretrained VLAs such as $\pi_{0.5}$~\cite{physicalintelligence2025pi05}, WAMs such as Fast-WAM~\cite{yuan2026fastwam}, and an imitation-learning baseline based on Diffusion Policy~\cite{chi2023diffusionpolicy}. Table~\ref{tab:main_results} lists all evaluated policies. Each policy is trained separately for atomic and compositional tasks, with model adaptations, training configurations, and preprocessing detailed in Appendix~B.

Policies use egocentric RGB observations and G1 proprioception where supported to predict action chunks for a compatible whole-body controller. Adapters connect policy inference to controller execution by handling observation processing, action representations, and normalization for each pairing. The main comparison uses SONIC motion tokens and finger targets, while the controller study trains policies for the respective controller interfaces. A shared driver delivers commands at 50~Hz, allowing policies to specify the intended behavior while controllers realize coordinated body motion. The supported controllers differ in how they represent these commands, ranging from base and arm targets to whole-body references and learned motion tokens.

The testbed integrates five whole-body controllers with complementary control strategies:
\begin{itemize}
  \setlength{\itemsep}{2pt}
  \setlength{\parsep}{0pt}
  \item AMO~\cite{li2025amo} coordinates learned locomotion and torso control with arm-motion control, allowing manipulation while maintaining balance.
  \item SONIC~\cite{luo2025sonic} learns a shared motion representation through large-scale motion tracking, supporting diverse whole-body behaviors with a unified control policy.
  \item HEFT~\cite{liu2026heft} learns robust whole-body tracking from noisy teleoperation references using privileged motion guidance and a payload curriculum.
  \item ALMI~\cite{shi2025almi} coordinates lower-body locomotion and upper-body motion imitation through adversarial training between the two policies.
  \item ScaleBFM~\cite{zeng2026scalebfm} scales whole-body motion tracking through diverse reference data and a Humanoid Transformer architecture.
\end{itemize}
Observations and commands are recorded at 50~Hz for all five controllers, with finger targets handled alongside the body-control interface.
The recorded state contains 43 measured joint positions for SONIC, HEFT, ALMI, and ScaleBFM.
AMO uses a 32-entry layout containing 28 measured arm-and-hand joints and four previous torso-command entries.
Appendix~B-B describes the policy-specific selection of recorded state fields and the corresponding controller command layouts.

\subsection{Task Design}

The task suite spans kitchen, living-room, bedroom, bathroom, and office scenes, with 13 scene instances and 234 instructed tasks.
Its 102 atomic tasks isolate skills including navigation, articulated-object interaction, picking, placing, and object reorientation.
The 132 compositional tasks combine two to four primitives into longer episodes, often linking manipulation through navigation.
Most compositional tasks combine atomic tasks, extending evaluation from individual skills to their coordination and long-horizon execution.
For the main benchmark, policies are trained separately on two 50-task lists, each covering all five scene categories.
Each checkpoint is evaluated on seen tasks, unseen tasks in training scenes, and unseen tasks in held-out scenes, testing both familiar-task execution and generalization to new tasks and environments.
Split assignments are defined separately for the atomic and compositional checkpoints, with task counts reported in Table~\ref{tab:main_results}.
Task success is determined by physical predicates, while stage annotations support demonstration collection and analysis.

\subsection{Dataset}
\label{sec:dataset}

\system{} combines three demonstration sources for policy learning and data-source comparisons.
Teleoperation captures human-guided execution in simulation and on the physical robot, while motion synthesis produces demonstrations by executing task-conditioned whole-body motions.
Trajectory augmentation adapts existing demonstrations to perturbed initial conditions and retains successful simulated executions as additional training data.
The teleoperated corpus includes 51.5 hours of simulation data across five controllers and approximately 9.0 hours of real-robot data, complemented by synthesized and augmented demonstrations for selected tasks.
Fig.~\ref{fig:data_stats} summarizes the simulated teleoperation corpus by scene, controller, trajectory duration, and body-motion patterns, characterizing its scale and behavioral diversity.

\subsubsection{Data Collection}

Human operators use a virtual-reality (VR) headset and hand controllers to teleoperate the humanoid through its whole-body controller, collecting task demonstrations in simulation and on the physical robot.
Each episode records controller commands, measured robot states, and egocentric RGB images, with image resolutions of $320\times240$ in simulation and $640\times480$ on hardware.
Simulation episodes end automatically when predefined task-specific success conditions are satisfied, while real-robot episodes are ended by the operator upon judging task completion.
At the start of each simulated episode, the initial horizontal positions of the robot and movable task objects are perturbed within 5~cm and 2~cm of their nominal locations, respectively.

\subsubsection{Data Generation}
\label{par:motion_synthesis}

\emph{Motion generator adaptation.}
The motion generation model Kimodo~\cite{rempe2026kimodo} is fine-tuned for task-oriented data generation.
As the motion diffusion model (MDM) in Fig.~\ref{fig:overview}, it generates whole-body motion chunks by iteratively denoising a motion sequence conditioned on action text, an initial motion prefix, and root or wrist targets.
Kimodo is paired with a frozen whole-body controller that executes the generated references during fine-tuning and data generation.
During fine-tuning, spatial commands specify a planar root-position target and a target end-effector (EEF) position represented at the wrist.
Fine-tuning combines randomly sampled spatial conditions with conditions extracted from benchmark task stages in equal proportions.
Random conditions sample palm goals around the robot and derive corresponding root and wrist targets.
Task-derived conditions obtain navigation goals and manipulation waypoints from task definitions and scene geometry, with navigation stages using root targets alone.
The denoiser is adapted using proximal policy optimization (PPO)~\cite{schulman2017ppo}, following diffusion-policy optimization~\cite{black2024ddpo}.
Each stochastic denoising transition defines a Gaussian distribution over the next motion estimate, allowing the sampled transition to be evaluated under the old and updated models.
The clipped objective uses per-transition probability ratios and advantages obtained by normalizing execution rewards across motions sampled under the same condition.
Rewards are computed from controller execution in an uncluttered simulator.
Palm-reaching rewards depend on sustained target proximity when achieved and terminal palm-target distance otherwise, while approach rewards use terminal root-target distance with torso and waist inclination penalties.
The adaptation study compares low-rank adaptation (LoRA)~\cite{hu2022lora} with full denoiser fine-tuning (Full), keeping the controller and non-denoiser Kimodo components frozen.

\emph{Scene-level generation.}
Task demonstrations are generated by pairing the fine-tuned Kimodo model with the frozen controller in a loop that alternates motion generation and physical execution.
Kimodo generates each motion chunk from current spatial targets and a prefix of the robot's executed motion, while the controller tracks the resulting reference motion in the scene.
The executed motion then conditions the next chunk, with reference smoothing at chunk boundaries to support continuous execution.

For navigation, A*~\cite{hart1968astar} plans a route to a standing region on an occupancy map that accounts for body and arm clearance, and the route is converted into root waypoints that preserve turns.
The route is replanned from the robot's current position when drift or stalls occur.
For manipulation, the pipeline uses object positions and shapes to construct a sequence of wrist targets that guide the hand toward the object, establish contact or a grasp, perform the required object motion, and release it.
Contact and grasp feedback determine when to advance through this sequence and when to close or open the fingers.
A language model can assist by selecting the operating hand and an interaction location from task and scene descriptions, after which geometric rules determine the wrist targets.
Generation uses hand choices determined in advance and wrist targets computed from object geometry or task-specific manual annotations, without querying a language model during rollout.
Successful rollouts are retained as demonstrations, with additional physical and motion-quality filtering in some collection stages.

\begin{figure*}[!t]
  \centering
  \includegraphics[width=\textwidth]{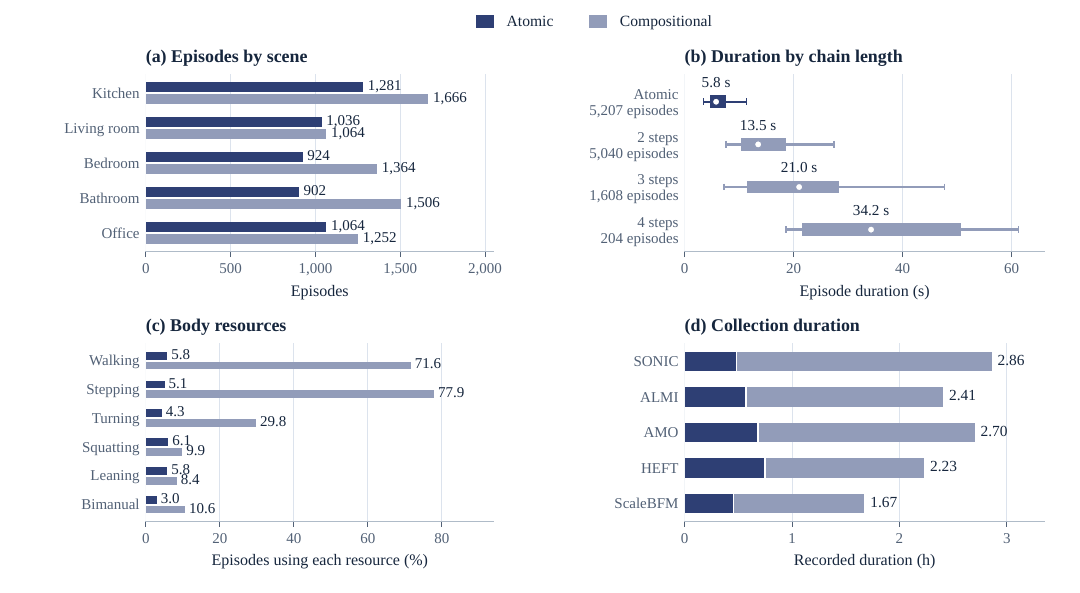}
  \caption{Teleoperation data statistics. Panels (a)--(c) describe the SONIC corpus by scene category, task-chain duration, and body motion. Duration is shown by median, interquartile range, and 5th--95th percentile. Panel (d) compares recorded execution time for 50 demonstrations per task on ten shared bathroom tasks, totaling 500 episodes per controller. Stacked segments separate atomic and compositional tasks, and bar labels give total hours.}
  \label{fig:data_stats}
\end{figure*}

\emph{Trajectory augmentation.}
Trajectory augmentation uses a single teleoperated demonstration as a motion prior for execution from perturbed initial conditions.
An optional large language model (LLM) planner receives task information, initialization offsets, and feedback from previous attempts. Its structured output selects wrist-target compensation, motion-generation text and keyframe spacing, and the allowed deviation from the demonstration.
A frozen controller executes the candidate motions, and task-completion and physical-validity checks determine which recordings are retained.
The data-source experiment uses a fixed augmentation routine that varies the timing of body and finger commands together and perturbs robot and object initial positions, without online LLM planning.

\subsubsection{Data Statistics}

Fig.~\ref{fig:data_stats} summarizes 14,160 simulated teleoperation episodes and 51.5 hours of motion across five controllers.
SONIC provides the largest collection, with atomic and compositional demonstrations across all five household scene categories.
The other four controllers contribute demonstrations on ten shared bathroom tasks for comparisons across control interfaces.
The SONIC recordings show increasing median episode duration and a wider spread of durations as task chains grow longer.
Atomic demonstrations primarily capture object interactions with little base movement, while compositional demonstrations more frequently involve walking, turning, and bimanual manipulation.
These motion patterns reflect the combination of navigation and object interaction in compositional tasks.
For the controller comparison in Fig.~\ref{fig:data_stats}(d), 50 demonstrations per task are selected for each controller on the same ten bathroom tasks.
At this equal episode count, shorter recorded durations indicate more efficient data collection.
The largest differences arise in compositional tasks, showing that the execution-time cost of collecting longer demonstrations varies across controllers.

\subsection{Randomization and Real-Robot Validation}
\label{par:randomization}

Domain randomization composes appearance, camera, and distractor perturbations separately from reset configuration.
Low changes textures and lighting. Mid adds camera and material-shading changes, while High also enables distractors, lens distortion, and robot-color changes.
Parameters can be overridden, so presets specify combinations rather than independent visual, geometric, and dynamics severity scales. Episode-derived seeds control sampling. Each DR episode applies the same reset randomization and seed as the corresponding episode without DR; when distractors are added, the robot and task objects are first returned to their distractor-free nominal placements before the randomization is applied.
These perturbations retain task definitions while changing visual context. They complement unseen task and scene evaluation by examining appearance sensitivity separately from performance on new activities and layouts.

Three real-world tasks are designed to evaluate policy performance on a Unitree G1 humanoid, with matching tasks in simulation used to examine whether simulation-based evaluation reflects policy performance on the physical robot.
The simulated tasks reproduce the corresponding activities and approximate the real-world scene layouts.
Policies use the same training and inference configurations in both settings, with real teleoperation replacing simulated demonstrations for real-world training.
Task completion on the physical robot is judged by a human.
The comparison examines consistency in policy performance across the two settings, while differences in scene geometry and success criteria limit direct comparisons of absolute success rates.

\section{Experiments}
\label{sec:experiments}
In this section, we evaluate the performance and generalization of humanoid loco-manipulation policies through systematic experiments on \system{} in simulation and the real world.
The main experiment compares twelve policies on seen and unseen tasks and scenes, followed by a robustness test under visual perturbations.
Studies of controller choice, demonstration scale, and data sources then examine how training and execution settings affect task success and motion quality.
Kimodo adaptation is evaluated through executed motion accuracy and task completion, and corresponding simulation and real-robot tasks provide a final comparison of policy performance across the two environments.

\subsection{Experimental Setup}
\label{sec:experimental_setup}

\subsubsection{Policies and training}
The main experiment evaluates the twelve policies introduced in Sec.~\ref{par:low_level}, including VLA and WAM models and the Diffusion Policy baseline.
All use the SONIC controller and are trained separately on an atomic task list and a compositional task list, each containing 50 tasks with ten from each scene category.
Appendix~A summarizes these datasets and their evaluation splits.

Each policy retains its original observation, preprocessing, action-chunk length, and optimization schedule.
Every configuration uses one training run, and the main benchmark evaluates the model obtained at the end of its training budget without validation-based or test-score-based selection.
The controller, demonstration scale, and data-source studies focus on GR00T N1.7 and $\pi_{0.5}$, with the task sets and training budgets specified in their respective subsections.

\subsubsection{Task splits}
The main benchmark evaluates 102 atomic and 132 compositional tasks.
Seen-task evaluation uses the training task suite and scenes with evaluation initial states, while unseen-task evaluation introduces new tasks in familiar scenes.
The unseen-scene split includes both new scenes and new task definitions, so it tests their combined effect.
Splits are defined separately for the atomic and compositional checkpoints, allowing a scene to be familiar to one checkpoint and unseen by the other.
Domain randomization (DR) tests robustness on seen tasks, while the controller, demonstration scale, and data-source studies evaluate the tasks and scenes used for their own training.

\subsubsection{Metrics}
\label{sec:metrics}
Success rate (SR) measures completion according to each task's recorded success predicate, with aggregation weights specified for each experiment.
The controller comparison also reports absolute mechanical output power, foot slip during contact, changes in body-joint targets, fall rate, and relative completion time.
SR uses all evaluation episodes, while the other five metrics describe successful episodes only. Slip and action-rate calculations additionally exclude invalid motion signals. Action rate measures the magnitude of changes in commanded body-joint targets per unit time, and relative time compares completion times against a task- and policy-specific reference shared across controllers. These measures therefore characterize how successful tasks are executed, although the set of solved tasks may differ between controllers.
Appendix~C gives the definitions, normalization, and aggregation rules.

\subsection{Main Results: Generalization and Domain Randomization}
\label{sec:main_bench}

\subsubsection{Seen and unseen evaluation}
Table~\ref{tab:main_results} compares all twelve policies without DR, using the same tasks for every policy.
Atomic and compositional columns follow the semantic task labels, while each task retains the checkpoint trained on its original task list.
Evaluation uses 20 trials per seen task and 10 per unseen task or scene.
Scores are calculated from recorded success counts and weight tasks equally, including when combining the separately trained atomic and compositional checkpoints.
Overall Avg. averages over all evaluated tasks across the three splits, and the rows are ordered by this score. The task counts in the table determine each split's contribution to the overall average.

\emph{Compositional tasks separate policies with similar atomic performance.}
DM0.5, XR-1, $\pi_{0.5}$, and GR00T N1.7 achieve similar success on seen atomic tasks, but XR-1 and DM0.5 have a larger advantage on compositional tasks.
SmolVLA and PRTS illustrate a different change in ordering. SmolVLA performs better on atomic tasks, while PRTS performs better on compositional tasks.
Atomic task rankings therefore do not fully describe performance when navigation and manipulation must be combined across several stages.
Together, atomic and compositional evaluations assess both the execution of individual skills and the coordination of navigation and manipulation across multiple stages.

\emph{High seen-task success leaves many unseen tasks unsolved.}
All policies lose substantial performance on unseen tasks.
Even XR-1, which leads both unseen atomic and compositional tasks, drops from 80.2\% on seen tasks to 25.2\% on unseen tasks. $\pi_{0.5}$ and GR00T N1.7 also decline sharply, and their advantage over InternVLA-A1.5 becomes much smaller on unseen tasks despite a clear gap on seen tasks. SmolVLA achieves a higher seen-task average than PRTS but falls below it on unseen tasks, showing that familiar-task rankings do not fully predict performance on new tasks. These results highlight the need to evaluate unseen-task generalization separately from proficiency on training tasks.

\emph{Generalization to unseen scenes remains limited.}
This split combines new task definitions with scenes excluded from training, requiring policies to handle changes in both task goals and the surrounding environment. Success remains low across all evaluated policies, with DM0.5 achieving the highest average of only 15.5\%. XR-1 and InternVLA-A1.5 show a similar difference between the two categories, whereas Diffusion Policy records no successes in either category. The relatively higher success on compositional tasks may reflect the richer training experience provided by compositional demonstrations. By combining navigation and manipulation, these demonstrations expose policies to more varied viewpoints, robot configurations, and intermediate interaction states, which may help them apply learned skills in unfamiliar scenes.

\begin{table*}[t]
  \caption{Success rate (\%) with SONIC and no DR. Averages weight tasks equally. Bold and underlined values mark the best and second best in each column.}
  \label{tab:main_results}
  \centering
  \footnotesize
  \setlength{\tabcolsep}{4pt}
  \begin{tabular}{@{}l*{10}{c}@{}}
    \toprule
    \multirow{2}{*}{Policy} & \multicolumn{3}{c}{Seen task} & \multicolumn{3}{c}{Unseen task} & \multicolumn{3}{c}{Unseen scene} & \multirow{2}{*}{\shortstack{Overall\\Avg.}} \\
    \cmidrule(lr){2-4}\cmidrule(lr){5-7}\cmidrule(lr){8-10}
    & Atom. & Comp. & Avg. & Atom. & Comp. & Avg. & Atom. & Comp. & Avg. & \\
    Tasks & 50 & 50 & 100 & 46 & 77 & 123 & 6 & 5 & 11 & 234 \\
    \midrule
    XR-1~\cite{team2026xiaomi} & \underline{85.2} & \textbf{75.1} & \textbf{80.2} & \textbf{23.7} & \textbf{26.1} & \textbf{25.2} & 0.0 & 18.0 & 8.2 & \textbf{47.9} \\
    DM0.5~\cite{dexmal2026dm05} & \textbf{90.1} & \underline{65.7} & \underline{77.9} & \underline{16.3} & \underline{16.8} & \underline{16.6} & 0.0 & \textbf{34.0} & \textbf{15.5} & \underline{42.7} \\
    $\pi_{0.5}$~\cite{physicalintelligence2025pi05} & 84.4 & 51.0 & 67.7 & 12.6 & 13.6 & 13.3 & \underline{1.7} & \underline{24.0} & \underline{11.8} & 36.5 \\
    GR00T N1.7~\cite{nvidia2026grootn17} & 82.3 & 54.1 & 68.2 & 11.3 & 11.7 & 11.5 & \underline{1.7} & 12.0 & 6.4 & 35.5 \\
    InternVLA-A1.5~\cite{ma2026internvla15} & 70.7 & 36.7 & 53.7 & 11.5 & 9.9 & 10.5 & 0.0 & 18.0 & 8.2 & 28.8 \\
    SmolVLA~\cite{shukor2025smolvla} & 73.9 & 17.4 & 45.7 & 6.7 & 2.5 & 4.1 & \underline{1.7} & 6.0 & 3.6 & 21.8 \\
    Fast-WAM~\cite{yuan2026fastwam} & 52.8 & 28.6 & 40.7 & 7.8 & 4.2 & 5.5 & \underline{1.7} & 4.0 & 2.7 & 20.4 \\
    PRTS~\cite{zhang2026prts} & 37.0 & 34.3 & 35.7 & 3.5 & 7.0 & 5.7 & \underline{1.7} & 4.0 & 2.7 & 18.4 \\
    $\Psi_0$~\cite{wei2026psi0} & 55.1 & 9.0 & 32.1 & 8.0 & 1.7 & 4.1 & 0.0 & 0.0 & 0.0 & 15.8 \\
    X-VLA~\cite{zheng2025xvla} & 54.5 & 6.0 & 30.3 & 5.2 & 2.9 & 3.7 & \underline{1.7} & 4.0 & 2.7 & 15.0 \\
    StarVLA-$\alpha$~\cite{ye2026starvlaalpha} & 47.6 & 13.5 & 30.6 & 6.7 & 1.2 & 3.3 & \textbf{5.0} & 0.0 & 2.7 & 14.9 \\
    Diffusion Policy~\cite{chi2023diffusionpolicy} & 27.6 & 1.8 & 14.7 & 4.1 & 0.0 & 1.5 & 0.0 & 0.0 & 0.0 & 7.1 \\
    \bottomrule
  \end{tabular}
\end{table*}

\subsubsection{Domain randomization on seen tasks}
\label{sec:randomization}
Table~\ref{tab:randomization} evaluates the same twelve policies with SONIC on seen tasks under Low, Mid, and High perturbations. Each policy performs 20 evaluation rollouts per task at each perturbation level.
As in Table~\ref{tab:main_results}, atomic and compositional categories follow the semantic task labels, while each task retains the checkpoint trained on its original task list.
The overall score weights tasks equally and averages the three DR levels equally.
The perturbation combinations are defined in Sec.~\ref{par:randomization} and Appendix~A-A, and the comparison measures robustness during evaluation rather than the effect of training with randomization.

\emph{Stronger perturbations reduce success, but policies differ in robustness.}
Average success across all evaluated policies decreases at each level in both task categories.
XR-1 leads both task categories at Low, but DM0.5 loses less performance as perturbations increase and ranks first in both categories at Mid and High.
On atomic tasks, $\pi_{0.5}$ also declines less than XR-1 and surpasses it at High.
A policy that performs best under mild perturbations therefore need not retain its lead under stronger changes.
The combined presets assess how well policies maintain task success under simultaneous changes in scene appearance, lighting, and camera conditions.

\begin{table}[t]
  \caption{Seen-task success rate (\%) across DR levels. Bold marks the best value and underlining marks the second best.}
  \label{tab:randomization}
  \centering
  \footnotesize
  \setlength{\tabcolsep}{3pt}
  \begin{tabular}{@{}l*{7}{c}@{}}
    \toprule
    \multirow{2}{*}{Policy} & \multicolumn{3}{c}{Atomic} & \multicolumn{3}{c}{Compositional} & \multirow{2}{*}{\shortstack{Overall\\Avg.}} \\
    \cmidrule(lr){2-4}\cmidrule(lr){5-7}
    & Low & Mid & High & Low & Mid & High & \\
    Tasks & 50 & 50 & 50 & 50 & 50 & 50 & 100 \\
    \midrule
    DM0.5~\cite{dexmal2026dm05} & \underline{89.3} & \textbf{86.9} & \textbf{83.0} & \underline{60.8} & \textbf{58.5} & \textbf{42.5} & \textbf{70.2} \\
    XR-1~\cite{team2026xiaomi} & \textbf{90.0} & \underline{85.4} & 70.5 & \textbf{71.1} & \underline{57.7} & \underline{39.5} & \underline{69.0} \\
    $\pi_{0.5}$~\cite{physicalintelligence2025pi05} & 83.5 & 81.7 & \underline{78.8} & 44.1 & 40.7 & 34.1 & 60.5 \\
    GR00T N1.7~\cite{nvidia2026grootn17} & 80.1 & 75.3 & 59.8 & 47.6 & 41.6 & 20.7 & 54.2 \\
    InternVLA-A1.5~\cite{ma2026internvla15} & 67.5 & 65.4 & 61.6 & 31.2 & 27.2 & 22.1 & 45.8 \\
    SmolVLA~\cite{shukor2025smolvla} & 68.5 & 66.9 & 58.0 & 15.7 & 13.5 & 8.9 & 38.6 \\
    $\Psi_0$~\cite{wei2026psi0} & 48.3 & 43.4 & 38.4 & 8.1 & 6.0 & 4.8 & 24.8 \\
    Fast-WAM~\cite{yuan2026fastwam} & 36.7 & 28.8 & 28.0 & 19.4 & 15.2 & 15.2 & 23.9 \\
    StarVLA-$\alpha$~\cite{ye2026starvlaalpha} & 41.2 & 32.7 & 25.4 & 11.0 & 9.0 & 3.0 & 20.4 \\
    PRTS~\cite{zhang2026prts} & 30.9 & 36.3 & 32.3 & 4.1 & 3.6 & 2.4 & 18.3 \\
    X-VLA~\cite{zheng2025xvla} & 39.4 & 32.7 & 27.9 & 2.1 & 2.4 & 1.9 & 17.7 \\
    Diffusion Policy~\cite{chi2023diffusionpolicy} & 10.0 & 10.5 & 9.7 & 0.2 & 0.1 & 0.0 & 5.1 \\
    \bottomrule
  \end{tabular}
\end{table}

\begin{figure*}[!t]
  \centering
  \includegraphics[width=\textwidth]{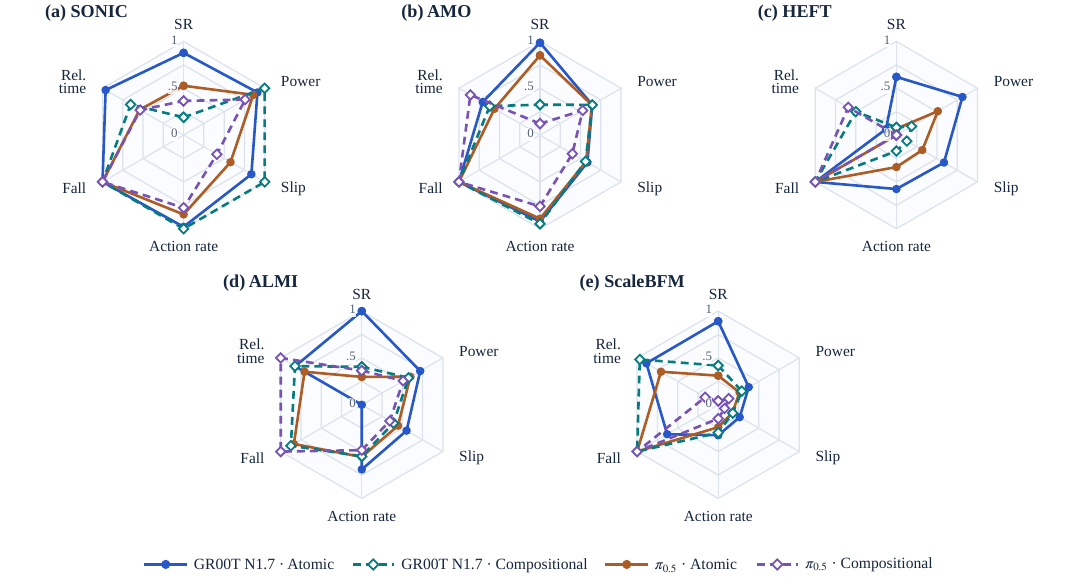}
  \caption{Controller comparison at 50 demonstrations per task. Each controller panel shows GR00T N1.7 and $\pi_{0.5}$ on atomic profiles with solid lines and filled circles, and compositional profiles with dashed lines and hollow diamonds. SR uses all episodes and other metrics use successful episodes only. Radial ticks show normalized values, with outward values better. Shared raw ranges, rounded to three decimals, are SR 5--79\%, power 13.795--317.734~W, slip 0.093--20.911~cm/s, action rate 3.048--45.392~rad/s, falls 0--22.785\%, and relative time 0.873--1.859. SR and falls use linear scales and other metrics use logarithmic scales.}
  \label{fig:controller_radar}
\end{figure*}

\subsection{Multi-Controller Evaluation}
\label{sec:controller_study}

The multi-controller evaluation compares GR00T N1.7 and $\pi_{0.5}$ with SONIC, AMO, HEFT, ALMI, and ScaleBFM on five atomic and five compositional bathroom tasks.
Each policy is trained for the corresponding controller's command representation, using separate atomic and compositional checkpoints and 50 demonstrations per task.
These are the same configurations evaluated at the shared 50-demonstration budget in the demonstration scale experiment.
Fig.~\ref{fig:controller_radar} shows task success and motion metrics for each policy, controller, and task category.
The metrics use shared normalization ranges across all profiles, with larger radii indicating better values on each axis.

\emph{The controller with the highest success need not produce the best motion.}
For GR00T N1.7 on atomic tasks, ALMI achieves the highest SR but has more falls among its successful episodes than SONIC.
SONIC uses the least mechanical output power during successful execution for both policies and task categories, yet other controllers achieve higher SR in several of these comparisons.
These results show that task completion alone is insufficient for choosing an execution controller, as configurations with comparable success can differ in power use, stability, and motion quality.
Since each policy is trained with controller-specific demonstrations and action targets, the comparison describes the resulting policy--controller pairs, including their training data and command representations.

\subsection{Demonstration Scale Study}
\label{sec:data_scale}

The demonstration scale study examines how the number of demonstrations used for post-training affects GR00T N1.7 and $\pi_{0.5}$ across all five controllers.
The task set is the same as in the multi-controller experiment, and atomic and compositional checkpoints are trained separately.
Fig.~\ref{fig:data_scale} compares 10, 30, and 50 demonstrations per task, using nested subsets shared by the two policies within each controller.
SONIC additionally includes 100 demonstrations per task, while the mean panel compares only the budgets available for all five controllers.

\par\medskip\noindent
\begin{minipage}{\linewidth}
  \centering
  \captionsetup{hypcap=false}
  \includegraphics[width=\linewidth]{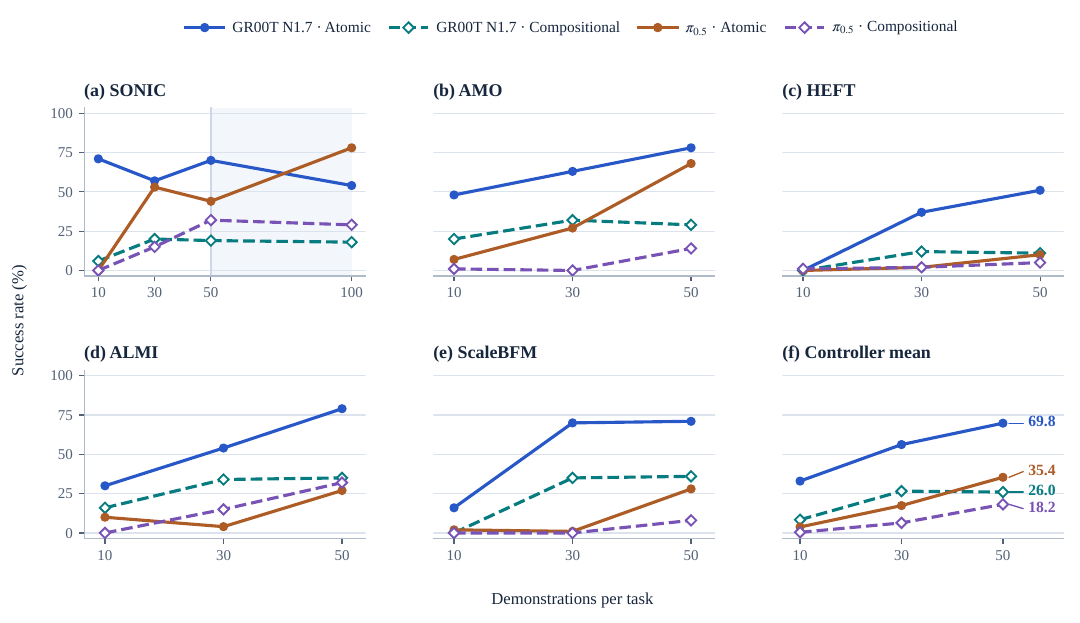}
  \captionof{figure}{Success rate versus demonstrations per task. Atomic and compositional checkpoints are trained separately, with 100 evaluation episodes per checkpoint. SONIC additionally includes 100 demonstrations per task, with shading highlighting this budget extension. The mean panel weights the five controllers equally at their shared budgets.}
  \label{fig:data_scale}
\end{minipage}
\par\medskip

Each checkpoint is evaluated on its five training tasks with 20 trials per task, and the mean panel gives each controller equal weight.
The 50-demonstration results also form the controller comparison in Fig.~\ref{fig:controller_radar}.
Training uses the model configurations of the main experiment in Sec.~\ref{sec:main_bench}, with the number of training updates increasing in proportion to the demonstration count.
Appendix~D-A specifies the training schedules and normalization settings.

\emph{More demonstrations and training improve most configurations.}
Success increases from 10 to 50 demonstrations per task for most combinations of policy, controller, and task category.
For example, GR00T N1.7's mean atomic SR rises from 33.0\% to 69.8\%, while $\pi_{0.5}$ improves on average in both task categories from a weaker starting point.

\emph{The benefit depends on the controller and task category.}
GR00T N1.7 improves substantially on atomic tasks with HEFT and ScaleBFM, whereas SONIC already performs well at the smallest budget and shows little improvement by 50 demonstrations.
Controller rankings therefore change with the available training data.
The atomic and compositional results also follow different trends. Between 30 and 50 demonstrations, GR00T N1.7's mean compositional SR levels off, with small decreases for three controllers, while its mean atomic SR continues to rise.
Over the same interval, $\pi_{0.5}$ improves on average in both categories.
Additional demonstrations and training therefore do not benefit all policy--controller pairs or both task categories equally. The SONIC extension shows that gains can also stop beyond the shared budget range.
Increasing the demonstration count from 50 to 100 per task further improves $\pi_{0.5}$ on atomic tasks, with a slight decrease in compositional success.

\subsection{Data Generation}
\label{sec:data_generation}

The data-generation experiments examine how synthesized and augmented demonstrations affect downstream policy performance and whether fine-tuning Kimodo improves motion generation and task success. The former is evaluated through mixtures of teleoperated, synthesized, and augmented demonstrations, while the latter compares the original generator with two fine-tuned variants.

\subsubsection{Data-source mixtures}
\label{sec:generated_data}
GR00T N1.7 and $\pi_{0.5}$ are trained with SONIC using six combinations of teleoperation, motion synthesis, and trajectory augmentation.
Each model is jointly trained and evaluated on the same 20 tasks across kitchen, living-room, bedroom, and bathroom scenes.
Table~\ref{tab:data_source} reports success rates for atomic and compositional tasks and an overall average that gives equal weight to each task.
The augmentation demonstration starts from one teleoperated trajectory per task and varies execution timing and initial conditions, retaining successful simulated executions.
Synthetic demonstrations are generated using fully fine-tuned Kimodo models and a frozen SONIC controller that executes the generated motions. Only rollouts that successfully complete the task are retained.

Mixtures concatenate the selected datasets without equalizing source counts or balancing their sampling probabilities.
GR00T N1.7 samples action-window starts and $\pi_{0.5}$ samples frame starts, giving longer trajectories more opportunities to contribute training samples.
The experiment therefore compares practical data combinations whose source proportions, total data volume, and training exposure differ.
Appendix~D-B and Appendix~D-C give the mixture settings and generation details.

\emph{Augmentation helps some tasks but does not improve every policy.}
Teleoperation is the strongest individual source for both models and gives their highest atomic success.
Adding augmentation raises GR00T N1.7's compositional SR from 56.2\% to 68.8\%, but slightly reduces its atomic SR, leaving the overall score nearly unchanged.
For $\pi_{0.5}$, teleoperation alone performs best in both categories, and none of the tested mixtures improves on it.
The benefit of augmentation is therefore specific to the evaluated model and task category.

Combining all three sources also fails to provide a consistent improvement.
GR00T N1.7 performs substantially worse with all three sources than with teleoperation plus augmentation, while $\pi_{0.5}$ achieves nearly the same aggregate success with the two mixtures.
Neither generated source alone matches teleoperation for either model.
Successful execution of generated demonstrations thus does not guarantee that adding them will improve the trained policy.

\begin{table}[t]
  \caption{Data-source success rate (\%) with SONIC. All averages weight tasks equally. Bold marks the best.}
  \label{tab:data_source}
  \centering
  \footnotesize
  \setlength{\tabcolsep}{3pt}
  \begin{tabular}{@{}l*{6}{c}@{}}
    \toprule
    \multirow{2}{*}{Training data} & \multicolumn{3}{c}{GR00T N1.7} & \multicolumn{3}{c}{$\pi_{0.5}$} \\
    \cmidrule(lr){2-4}\cmidrule(lr){5-7}
    & Atom. & Comp. & All & Atom. & Comp. & All \\
    Tasks & 16 & 4 & 20 & 16 & 4 & 20 \\
    \midrule
    Teleoperation only & \textbf{88.4} & 56.2 & 82.0 & \textbf{92.5} & \textbf{76.2} & \textbf{89.2} \\
    Motion synthesis only & 40.9 & 22.5 & 37.2 & 46.9 & 20.0 & 41.5 \\
    Traj. augmentation only & 66.9 & 23.8 & 58.2 & 63.7 & 3.8 & 51.7 \\
    Teleop. + synthesis & 73.8 & 63.7 & 71.8 & 88.1 & 73.8 & 85.2 \\
    Teleop. + augmentation & 85.9 & \textbf{68.8} & \textbf{82.5} & 90.3 & 63.7 & 85.0 \\
    All three & 46.9 & 43.8 & 46.2 & 89.7 & 63.7 & 84.5 \\
    \bottomrule
  \end{tabular}
\end{table}

\begin{table*}[t]
  \caption{Kimodo task success and goal error. Bold marks the best rates and mean errors.}
  \label{tab:kimodo_ft}
  \centering
  \begin{minipage}[t]{0.46\textwidth}
    \centering
    \footnotesize
    \textbf{(a) Task success (\%)}\par\smallskip
    \setlength{\tabcolsep}{4pt}
    \renewcommand{\arraystretch}{1.2}
    \begin{tabular}{@{}lccc@{}}
      \toprule
      Task group & Pretrained & LoRA & Full \\
      \midrule
    Overall & 39.7 & 71.3 & \textbf{90.7} \\
    Approach & 9.0 & 63.0 & \textbf{99.0} \\
    Comp. approach & 2.0 & 80.0 & \textbf{100.0} \\
    Close & 72.7 & 74.0 & \textbf{82.0} \\
      \bottomrule
    \end{tabular}
  \end{minipage}\hfill
  \begin{minipage}[t]{0.52\textwidth}
    \centering
    \footnotesize
    \textbf{(b) Goal error (m) $\downarrow$ (mean $\pm$ standard deviation)}\par\smallskip
    \newcommand{\kimodosd}[1]{\,\ensuremath{\pm}\,#1}
    \renewcommand{\arraystretch}{1}
    \setlength{\tabcolsep}{2pt}
    \begin{tabular}{@{}llccc@{}}
      \toprule
      Condition & Target & Pretrained & LoRA & Full \\
      \midrule
    Loco & Root & 0.693\kimodosd{0.286} & 0.096\kimodosd{0.055} & \textbf{0.076}\kimodosd{0.033} \\
    Mani & Root & \textbf{0.014}\kimodosd{0.008} & 0.036\kimodosd{0.015} & 0.051\kimodosd{0.015} \\
    Mani & Palm & 0.168\kimodosd{0.049} & 0.128\kimodosd{0.033} & \textbf{0.077}\kimodosd{0.037} \\
    Loco+Mani & Root & 0.578\kimodosd{0.150} & 0.089\kimodosd{0.052} & \textbf{0.080}\kimodosd{0.033} \\
    Loco+Mani & Palm & 0.650\kimodosd{0.147} & 0.122\kimodosd{0.073} & \textbf{0.099}\kimodosd{0.042} \\
      \bottomrule
    \end{tabular}
  \end{minipage}
\end{table*}

\subsubsection{Kimodo: pretrained, full fine-tuning, and LoRA}
\label{sec:kimodo_ft}
This experiment compares the original model with full denoiser fine-tuning, denoted Full, and LoRA adaptation, using the same frozen SONIC controller.
Both adaptation methods use an equal mix of randomly sampled and task-derived spatial commands. Random commands sample palm goals around the robot and derive corresponding root and wrist targets, while task-derived commands obtain navigation goals and manipulation waypoints from benchmark task definitions and scene geometry. Navigation stages use root targets alone, and each task stage is executed and scored independently during fine-tuning.
Each adapted variant uses one training run, and the model with the highest validation reward among saved checkpoints available at selection time is selected independently of the reported test results.

\emph{Fine-tuning improves task success most strongly on approach tasks.}
Table~\ref{tab:kimodo_ft}(a) reports results on 30 task--scene pairs with ten trials per pair and model.
Completion follows each task's success predicate, with the evaluation conditions detailed in Appendix~D-D.
Both fine-tuning methods improve success across the three reported task groups. Full fine-tuning raises overall SR from 39.7\% to 90.7\% and outperforms LoRA.
The largest gains occur in Approach and compositional Approach, with smaller improvements in Close.

\emph{More accurate executed motions accompany the gains in task success.}
{\looseness=-1\widowpenalty=10000
Table~\ref{tab:kimodo_ft}(b) measures root and palm goal errors after three seconds of controller execution under locomotion, stationary manipulation, and combined conditions.
Root error is the planar distance to the root target, while palm error is the three-dimensional distance from the executed grasp center to its target.
Models share prompts, motion prefixes, targets, and generation seeds and their physical executions start from the first poses of their own generated references.
The target ranges overlap those used for adaptation, so this test measures accuracy under related conditions.
Errors are reported in meters as means and standard deviations across conditions, with the full protocol in Appendix~D-D.\par}

Full fine-tuning achieves the lowest mean error in four of the five reported measures.
In the stationary manipulation test, the root target is fixed at the robot's initial position, so a lower root error indicates less base movement.
Small base adjustments are nevertheless acceptable when they help the hand reach its target.
The original model keeps its base closest to the initial position but has the largest palm error, whereas both fine-tuned models improve palm accuracy, with Full achieving the largest improvement.
Full's slightly higher root error should therefore be considered together with its lower palm error, since keeping the base fixed is not sufficient to complete the reaching motion.
Under combined locomotion and manipulation, Full reduces both root and palm errors, bringing the robot's body and hand closer to their respective targets.
These improvements are consistent with the higher success rates on approach and compositional approach tasks, and Fig.~\ref{fig:kimodo_ft} in Appendix~\ref{app:generation} illustrates the corresponding motion differences.

\begin{table*}[t]
  \caption{Simulation and Unitree G1 real-robot success rates (\%) on three tasks. Hardware results use ten trials per model--task pair. Bold marks the best value and underlining marks the second best.}
  \label{tab:real_robot}
  \centering
  \footnotesize
  \setlength{\tabcolsep}{7pt}
  \begin{tabular}{@{}l*{8}{c}@{}}
    \toprule
    \multirow{2}{*}{Policy} & \multicolumn{2}{c}{Fridge} & \multicolumn{2}{c}{Cup} & \multicolumn{2}{c}{Conveyor} & \multicolumn{2}{c}{Average} \\
    \cmidrule(lr){2-3}\cmidrule(lr){4-5}\cmidrule(lr){6-7}\cmidrule(lr){8-9}
    & Sim. & Real & Sim. & Real & Sim. & Real & Sim. & Real \\
    \midrule
    XR-1~\cite{team2026xiaomi} & \textbf{100.0} & \textbf{80.0} & \textbf{85.0} & \textbf{70.0} & \textbf{45.0} & \textbf{60.0} & \textbf{76.7} & \textbf{70.0} \\
    GR00T N1.7~\cite{nvidia2026grootn17} & \underline{55.0} & \underline{50.0} & \underline{80.0} & \underline{60.0} & \underline{25.0} & \underline{50.0} & \underline{53.3} & \underline{53.3} \\
    Fast-WAM~\cite{yuan2026fastwam} & 45.0 & 40.0 & \textbf{85.0} & 50.0 & 20.0 & \underline{50.0} & 50.0 & 46.7 \\
    \bottomrule
  \end{tabular}
  \par\smallskip
\end{table*}

\begin{figure*}[!t]
  \centering
  \includegraphics[width=\textwidth]{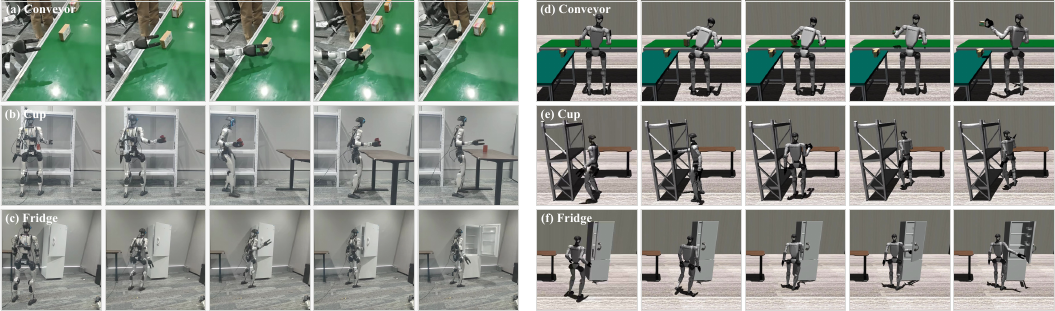}
  \caption{Task-execution sequences in real-world (a--c) and simulated (d--f) settings, arranged as real-world sequences in the left column and simulated sequences in the right column. Rows show conveyor box handling, cup transport, and fridge opening. Selected frames progress from left to right within each row. Real and simulated sequences are not temporally aligned, and success is measured separately in each environment. Corresponding rates appear in Table~\ref{tab:real_robot}.}
  \label{fig:real2sim}
\end{figure*}

\subsection{Simulation and Real-Robot Evaluation}
\label{sec:real_robot}

GR00T N1.7, Fast-WAM, and XR-1 are evaluated on fridge opening, cup transport from a shelf to a desk, and conveyor box handling in corresponding simulated and real-world scenes. Each policy is trained separately for each task, using simulated teleoperation for simulation and 50 real-robot demonstrations per task for hardware. Training uses the model configurations of the main experiment in Sec.~\ref{sec:main_bench}, with SONIC used for execution; training budgets are given in Appendix~E.
Evaluation uses 20 simulation episodes and 10 hardware trials per task and policy. Table~\ref{tab:real_robot} reports success rates, Fig.~\ref{fig:real2sim} shows execution examples, and Appendix~E details the task criteria and protocols.

XR-1 achieves the highest average success in both environments and leads all three hardware tasks. 
GR00T N1.7 ranks second and Fast-WAM third in both simulation and hardware, giving a consistent overall policy ranking that supports the use of simulation to assess relative real-world performance. Task-level results provide a more detailed comparison. GR00T N1.7 and Fast-WAM perform well on cup transport in simulation, but their ordering on this task reverses on hardware.
Conveyor handling has the lowest simulation success for all three policies, while GR00T N1.7 and Fast-WAM achieve the same hardware success on this task. Simulation therefore captures the overall differences between policies, while individual tasks reveal how their relative strengths vary across the two environments.

On hardware, fridge opening produces the largest gap between XR-1 and the other policies, whereas success rates are closer on cup transport and conveyor handling.
The stronger separation on fridge opening highlights articulated-object interaction as a useful complement to object transport and moving-object handling when comparing loco-manipulation policies.

\section{Discussion and Limitations}

Although most policies perform well on familiar tasks, unseen tasks and scenes leave substantial room for improvement. 
Policies with similar performance on atomic tasks can differ substantially when navigation and manipulation must be coordinated in compositional tasks.
Controller choice also matters, since comparable task success can be accompanied by different power use and stability. 
Together, these results highlight the value of testing policies across task complexity, scene variation, and execution controllers. 
Fine-tuning Kimodo improves both reaching accuracy and task completion, while the benefits of additional demonstrations and mixed data sources depend on the policy and task.
Task success generally improves with larger demonstration budgets, with gains varying across policies, controllers, and task categories.

These findings should be considered alongside limitations in evaluation coverage and deployment conditions. The benchmark primarily covers indoor household and office tasks. Simulation and physical trials yield the same ranking by average task success, supporting simulated comparisons before physical deployment. However, the hardware study covers only three policies on fridge opening, cup transport, and conveyor box handling, with ten trials per policy and task. 
Broader coverage of models, activities, and environments, together with repeated training and more physical trials, would help determine how consistently these trends extend to other settings. 
Deployment on robots with limited onboard compute may also require inference on nearby edge servers or cloud resources~\cite{shao2025ai,an2026ai}, making communication delays and missed control deadlines additional factors to evaluate.

\section{Conclusion}

\system{} provides a unified testbed for humanoid loco-manipulation, connecting household tasks, teleoperated and generated demonstrations, policy training, and evaluation in simulation and the real world.
The experiments examine policy generalization, controller performance, demonstration scale, data sources, and motion generator fine-tuning.
Although most policies perform well on familiar tasks, substantial room for improvement remains on unseen tasks and in unseen scenes, and controller choice affects both task completion and motion quality.
The data collection and generation pipelines support policy training with teleoperated, synthesized, and augmented demonstrations, enabling systematic study of how training data affects humanoid task performance.
Simulation and hardware yield the same average policy ordering on the evaluated tasks, supporting the use of simulated tasks to guide policy selection for physical evaluation.

\bibliographystyle{plainnat}
\bibliography{references}

\appendices
\section{Task Suite and Evaluation Protocol}
\label{app:tasks}

The collection contains 14,160 simulated teleoperation episodes (51.5 h) across five household scene categories. The principal SONIC corpus contains 5,207 atomic and 6,852 compositional episodes, and four other controllers contribute 2,101 episodes on ten shared bathroom tasks. Main-benchmark training uses separate 50-task atomic and compositional lists (2,546 and 2,558 episodes, respectively). The task suite comprises 102 atomic and 132 compositional tasks (Table~\ref{tab:tro_task_splits}).

\par\medskip\noindent
\begin{minipage}{\linewidth}
  \captionsetup{hypcap=false}
  \captionof{table}{Task inventory and evaluation splits. Split membership is defined separately for each atomic or compositional checkpoint; atomic and compositional counts follow the semantic task labels.}
  \label{tab:tro_task_splits}
  \centering\footnotesize
  \setlength{\tabcolsep}{5pt}
  \begin{tabular}{@{}lrrrr@{}}
    \toprule
    Scene category & Scene instances & Atomic tasks & Compositional tasks & Representative objectives \\
    \midrule
    Kitchen & 1 & 25 & 33 & articulated appliances; object placement \\
    Living room & 2 & 20 & 20 & navigation; object transport and reorientation \\
    Bedroom & 4 & 18 & 26 & drawers; navigation followed by interaction \\
    Bathroom & 4 & 18 & 28 & lids, shower fixtures, and multi-stage tasks \\
    Office & 2 & 21 & 25 & drawers, chairs, desk-object interaction \\
    \midrule
    Evaluated total & 13 & 102 & 132 & 234 task IDs \\
    \midrule
    Split & Atomic tasks & Compositional tasks & Episodes/task & Novelty \\
    Seen & 50 & 50 & 20 & training definitions and scenes \\
    Unseen task & 46 & 77 & 10 & unseen definitions, training scenes \\
    Unseen scene & 6 & 5 & 10 & unseen definitions and scene instances \\
    \bottomrule
  \end{tabular}
\end{minipage}
\par\medskip

Atomic objectives may contain several motion phases, whereas compositional objectives require ordered stages whose later actions depend on earlier state changes. Success follows task-specific physical predicates and any configured hold interval.

\begin{figure*}[p]
  \centering
  \includegraphics[width=\textwidth]{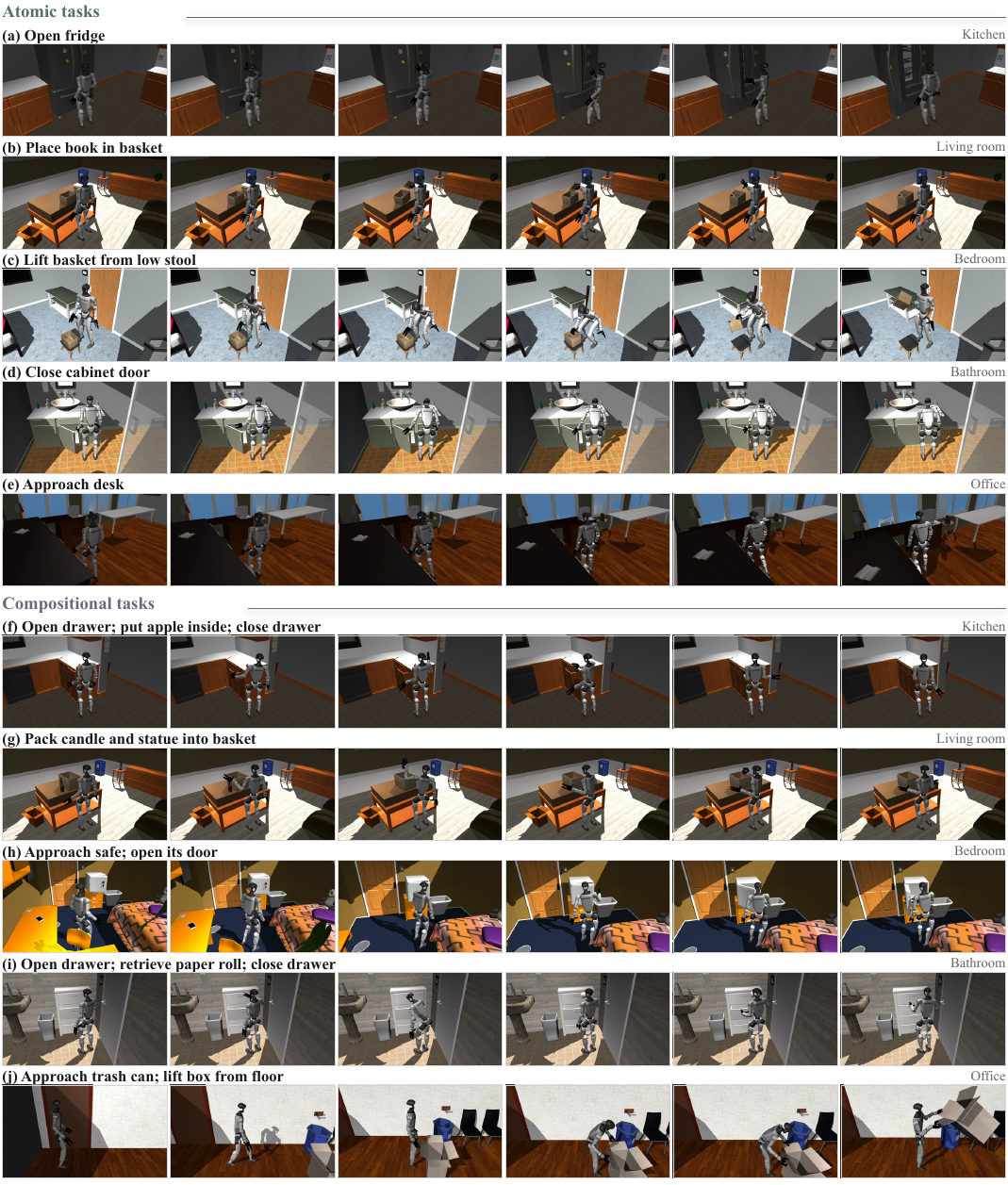}
  \caption{Representative task demonstrations across five scene categories. Upper rows show atomic objectives: (a) open a fridge, (b) place a book in a basket, (c) lift a basket from a stool, (d) close a cabinet, and (e) approach a desk. Lower rows show compositional objectives: (f) open a drawer, place an apple inside, and close it; (g) pack a candle and statue into a basket; (h) approach and open a safe; (i) open a drawer, retrieve a paper roll, and close it; and (j) approach a trash can and lift a box. Each row contains six chronological frames from a recorded SONIC demonstration, rendered through action replay; nonuniform frame intervals highlight task stages.}
  \label{fig:tro_task_sequences}
\end{figure*}

\begin{figure*}[t]
  \centering
  \includegraphics[width=\textwidth]{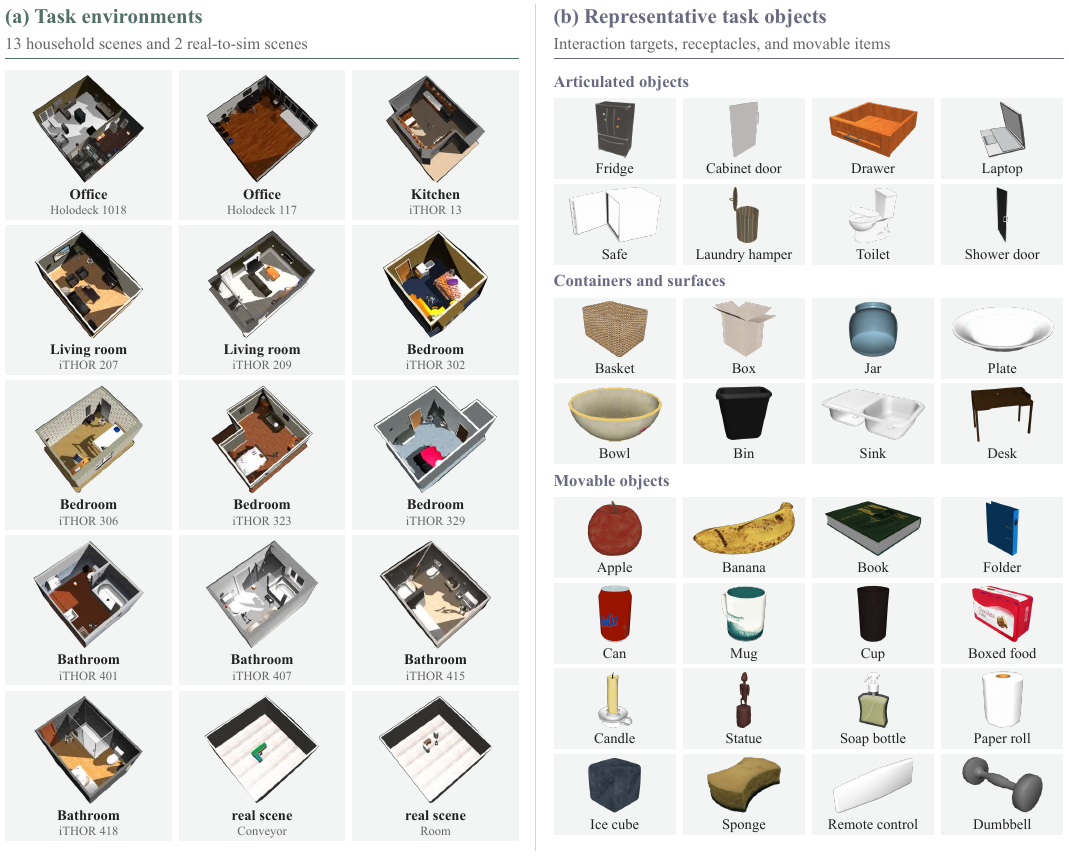}
  \caption{Task environments and representative interaction objects. (a) Thirteen household scenes from Holodeck and iTHOR, labeled by room category and scene identifier, followed by two Lab layouts. (b) Thirty-two representative task assets, grouped into articulated objects, containers and surfaces, and movable objects. The groups illustrate affordances for opening and closing, object placement, and manipulation within atomic and compositional tasks. Rooms and objects are individually scaled for visibility and are not shown to a common physical scale.}
  \label{fig:tro_rooms_objects}
\end{figure*}

Fig.~\ref{fig:tro_task_sequences} and Fig.~\ref{fig:tro_rooms_objects} show representative task demonstrations and the scenes and objects used in the suite.


The principal evaluation uses a Unitree G1, one $320\times240$ head RGB image and model-specific proprioception, SONIC at 50~Hz, and synchronous policy queries every 20 control steps. Only the leading actions of longer chunks are executed. Shorter chunks hold their last action until the next query. Episode seeds are 0--19 for seen tasks and 0--9 for unseen tasks, with scene seed zero and robot/object reset randomization up to 5/2~cm. A task's rollout limit is $\max\{\operatorname{round}(2\bar L),L_{\max},300\}$ control steps, where $\bar L$ and $L_{\max}$ are its mean and maximum demonstration lengths. Episodes end on success or timeout, without fall-based early termination.

\subsection{Domain Randomization Presets}
\label{app:dr}
Low/Mid/High replace 50/70/70\% of eligible texture groups from an unseen texture pool. All levels perturb lighting color and intensity by 0.15, toggle shadows with probability 0.5, and add a spotlight with probability 0.5. Mid adds material shading randomization of 0.2 and camera offsets of $\pm0.01$~m, $\pm0.05$~rad, and $\pm2^\circ$ field of view. High adds radial distortion with $k_1\sim U(-0.25,0.25)$ and $k_2\sim U(-0.05,0.05)$, principal-point shifts up to 8 pixels, image zoom up to 1.12, and robot RGB perturbations up to 0.15. High also places up to three distractors where space permits, which occurred in 38 of the 100 tasks.

\section{Policy Adaptation and Training}
\label{app:policies}

\subsection{Main-Benchmark Configuration}
\label{app:main_training}

The twelve evaluated policies are trained separately on the atomic and compositional 50-task lists. Inputs are a $320\times240$ head RGB image and 43 joint-state values, with three additional gravity values for GR00T N1.7 and no proprioception for StarVLA-$\alpha$. All adapters output 64 SONIC motion-token values and 14 absolute finger targets.

Budgets were set to $1{,}000\lceil 2U/(1{,}000B)\rceil$ updates for estimated sampling-unit count $U$ and global batch $B$. Most models use frame counts and GR00T N1.7 uses complete 64-step action-window starts. These correspond to approximately two nominal passes over model-specific sampling units and do not imply equal compute budgets.

Below, state/action normalization is denoted Q for training-set 1st/99th percentiles mapped to $[-1,1]$, M for mean/standard deviation, R for min/max scaling, and I for identity. Statistics are fitted on each training selection. Paired training budgets are reported in atomic/compositional order.

\noindent\textbf{Diffusion Policy.}
The RoboCasa365 implementation uses Contrastive Language--Image Pretraining (CLIP) features from a Vision Transformer (ViT-L/14) and a frozen text encoder. Two observations condition a 32-step prediction that returns 16 actions, with R/R normalization and 100 denoising steps using $\epsilon$ prediction and a squared-cosine noise schedule. Training uses global batch 192, AdamW at $10^{-4}$, cosine decay, 1k warmup updates, no gradient clipping, and seed 42. The budgets are 9k/26k optimizer updates. Evaluation uses exponential moving average (EMA) weights. Images use random $216\times288$ crops during training and center crops at test.

\noindent\textbf{Fast-WAM.}
Wan2.2 TI2V-5B and a pretrained action DiT are adapted while the Text-to-Text Transfer Transformer (T5) text encoder and the variational autoencoder (VAE) remain frozen. Each sample spans 33 time steps at 10~Hz over 3.2~s, supervising 32 actions and sampling every fourth time step for nine video frames including the conditioning frame. The action chunk is interpolated to 160 steps at 50~Hz. Training uses R/R normalization, global batch 64, cosine decay from $10^{-4}$, weight decay 0.01, bfloat16 (bf16), seed 42, and 25k/78k updates. Inference uses ten denoising steps. Instructions are wrapped in the training template ``A video recorded from a robot's point of view executing the following instruction: \{task\}'' and encoded by T5.

\noindent\textbf{GR00T N1.7.}
Initialization uses the released GR00T-N1.7-3B policy with a Cosmos-Reason2-2B vision-language backbone. State and action are padded internally to 132 dimensions, with Q/Q normalization and clipping. The 64-step action horizon uses four flow-inference steps. Training updates the projector and action-diffusion module while freezing the vision and language encoders, with global batch 128, AdamW, cosine decay from $10^{-4}$, weight decay $10^{-5}$, clipping at 0.8, seed 42, and 10k/37k updates. Images are square-padded, resized to $256\times256$, cropped to 95\% randomly in training or centrally at test, and resized back to $256\times256$. Training adds color randomization and state dropout with probability 0.2.

\noindent\textbf{InternVLA-A1.5.}
The adapted policy combines Qwen3.5-2B with a 50-step action expert. The data pipeline applies M/M normalization using training-set statistics. Action-only loss updates the entire vision-language model (VLM), including its vision and language components, and the action expert while freezing the WAN video DiT. Training uses global batch 32, AdamW with peak rate $5\times10^{-5}$, cosine decay, 2k warmup updates, weight decay 0.01, seed 42, and 50k/156k updates. Inference uses ten flow steps and seed 42.

\noindent\textbf{$\pi_{0.5}$.}
OpenPI uses a 16-step action horizon, an 80-dimensional internal action layer sliced to 78 outputs, and unclipped Q/Q normalization. Full fine-tuning uses global batch 32, AdamW with clipping at 1.0, seed 42, and 50k/156k updates. The learning rate is constant at $5\times10^{-5}$ after 1k warmup updates. EMA decay is 0.999, and EMA weights are evaluated with ten flow steps. Training augments images with random 95\% crops, rotations up to $\pm5^\circ$, and color randomization.

\noindent\textbf{PRTS.}
PRTS-4B/Qwen3-VL is fully fine-tuned, with a 20-step chunk and Q/Q normalization. Training uses global batch 16, peak rate $5\times10^{-5}$, 5k warmup updates, linear decay, seed 42, and 99k/312k updates. Up to three randomly selected image transforms vary color, sharpness, crop, or rotation. Inference uses five flow steps.

\noindent\textbf{$\Psi_0$.}
The frozen VLM conditions a trainable action expert with a 30-step horizon and R/R normalization. Training uses global batch 128, bf16, AdamW at $10^{-4}$, cosine decay, 1k warmup updates, seed 292285, and 13k/39k updates, with color randomization and Gaussian image noise. Inference uses eight flow steps.

\noindent\textbf{SmolVLA.}
The SmolVLM2 backbone, including vision and language components, is frozen while the action expert and state projection are trained. The adapted policy uses a 50-step chunk, M/M normalization, and ten flow-inference steps. Training uses global batch 128, AdamW, seed 1000, and 13k/39k updates, with the learning-rate schedule shared with X-VLA described below.

\noindent\textbf{StarVLA-$\alpha$.}
Qwen3-VL-4B-Instruct-Action uses QwenOFT full fine-tuning, no proprioception, R action normalization, and 16-step action regression without stochastic sampling. Training uses global batch 64, base/interface and action learning rates of $10^{-5}$ and $10^{-4}$, cosine decay, 5k warmup updates, seed 42, and 25k/78k updates.

\noindent\textbf{XR-1.}
Xiaomi-Robotics-1 is adapted to the SONIC interface by padding state to 60 dimensions and slicing 80 action outputs to 78 over a 30-step chunk. State uses Q normalization, and actions use M statistics for each chunk position. All components except VLM input embeddings are trained, with global batch 32, seed 42, and 50k/156k updates. The scheduler warms from $5\times10^{-7}$ to $2\times10^{-5}$ over 500 updates, then decays to $5\times10^{-6}$. Training varies brightness, contrast, and saturation. Inference uses five Euler flow steps.

\noindent\textbf{X-VLA.}
The adapted policy uses I/M normalization, a 30-step chunk, and ten inference steps. Fine-tuning updates the transformer and soft prompts without freezing vision or language modules, using global batch 128, AdamW, seed 1000, and 13k/39k updates. Only the egocentric view is available; the two unused camera slots are masked in both training and evaluation.

\noindent\textbf{DM0.5.}
The released DM0.5 base pairs a Gemma~3 vision-language model, whose SigLIP encoder reads $448\times448$ images, with a 34-layer flow-matching action expert. Its action input and output projections are widened from 32 to 78 dimensions. The 43 joint-state values enter the prompt as 256-bin state tokens, and actions use OpenDM's absolute rather than state-relative mode over a 50-step chunk, with Q/Q normalization. Full fine-tuning uses global batch 32, AdamW with clipping at 1.0, peak rate $2.5\times10^{-5}$, 1k warmup updates, cosine decay to 10\% of the peak, seed 42, and 50k/156k updates. Images are padded to square and resized to $448\times448$; training applies random 95\% crops, rotations up to $\pm2^\circ$, and color randomization, each with probability 0.5. Inference uses ten flow steps.

SmolVLA and X-VLA use a cosine schedule with peak $10^{-4}$ and floor $2.5\times10^{-6}$: atomic runs warm up for 433 updates and decay over 13k updates, while compositional runs warm up for 1k updates, decay over 30k updates, and stay at the floor for the last 9k. For image preprocessing, $\pi_{0.5}$, InternVLA-A1.5, and X-VLA pad images to $224\times224$, and SmolVLA to $512\times512$. InternVLA-A1.5, SmolVLA, StarVLA-$\alpha$, X-VLA, and Fast-WAM use no training image augmentation.

\subsection{Controller-Specific Policy Interfaces}
\label{app:controller_interfaces}
The five controller interfaces expose 36 (AMO), 78 (SONIC), 50 (HEFT), 52 (ALMI), and 50 (ScaleBFM) action values at 50~Hz. AMO combines 22 body-command entries with 14 fingers; SONIC uses 64 motion tokens and 14 fingers; HEFT and ScaleBFM use root pose plus 29 joints and 14 fingers; ALMI combines four base commands and 34 upper-body values with 14 fingers. The controller comparison uses GR00T N1.7 and $\pi_{0.5}$, with policies trained on the command layout and data associated with each controller.

Both policies receive an egocentric RGB image and a task instruction, while their state inputs depend on the controller. With SONIC, HEFT, ALMI, and ScaleBFM, $\pi_{0.5}$ uses the 43 measured body-and-hand joint positions, and GR00T N1.7 additionally uses three projected-gravity values, giving 46 state inputs. With AMO, both policies use 32 entries comprising 28 measured arm-and-hand joint positions and the previous torso orientation and height commands, without projected gravity. These fields are used consistently during training and inference, and the dimensions refer to selected inputs before internal model padding. Predicted actions follow the corresponding controller command layout above, including the 14 finger targets, and are delivered at 50~Hz.

\section{Metrics and Aggregation}
\label{app:metrics}

For task $t$, with $n_t$ successes among $N_t$ trials, success rate is averaged over tasks rather than pooled episodes:
\begin{equation}
  \operatorname{SR}(\mathcal T)=\frac{100}{|\mathcal T|}\sum_{t\in\mathcal T}\frac{n_t}{N_t}.
\end{equation}
Overall averages use the task weights stated in each table. The DR aggregate weights its 100 tasks and three levels equally. For Fig.~\ref{fig:controller_radar}, SR uses all 100 episodes per profile. Power, slip, action rate, falls, and relative time use successful episodes only, with 5--79 successes per profile. Each is calculated per episode and then averaged equally across successful episodes in the profile, rather than averaging task-level means.

Mechanical output energy integrates absolute motor power over 43 simulated motors, $E_e=\sum_k\Delta t\sum_j|\tau_{e,j,k}\dot q_{e,j,k}|$, with a 2~ms physics step in controller studies. Reported power is $E_e/T_e$ in watts, where $T_e$ is the episode duration; the power omits electrical losses. Foot slip is the mean horizontal foot-body speed in cm/s over samples with contact between a foot and any non-robot geometry. Action rate is the mean norm of successive differences in the 29 body-joint position targets, divided by the nominal 20~ms control interval. Fall rate flags a decrease in free-joint root height exceeding 0.30~m from its first recorded value.

Relative time divides each successful episode's duration by the median successful duration for the same task and policy. The reference pool combines successful episodes across all five controllers at 10, 30, and 50 demonstrations per task, together with SONIC's 100-demonstration setting. Its median is computed over episodes rather than controller-level means or medians. This common denominator is used for all controllers of a given policy and task, including the 50-demonstration profiles in Fig.~\ref{fig:controller_radar}. Relative time is therefore model-specific and does not compare absolute speed between models. Unsuccessful episodes are excluded when computing power, energy, and relative time.

Radar bounds $L,U$ are each metric's minimum and maximum across the 20 profiles, with raw ranges in Fig.~\ref{fig:controller_radar}. Define $z=[g(x)-g(L)]/[g(U)-g(L)]$, using identity $g$ for SR and falls and $g(x)=\ln x$ otherwise. The radius is $z$ for SR and $1-z$ for the other metrics. Shared scales preserve relative ordering but polygon area is not an aggregate score.

\section{Auxiliary Data and Generator Studies}
\label{app:auxiliary}

\subsection{Controller and Demonstration Scale Protocol}
The controller and demonstration scale studies use five atomic and five compositional bathroom tasks in scenes 401, 415, and 418. Nested subsets contain $N\in\{10,30,50\}$ demonstrations per task, with $N=100$ additionally available for SONIC. Separate five-task models use $100N$ updates on $5N$ trajectories, with global batch 64 for GR00T N1.7 and 32 for $\pi_{0.5}$. Each configuration has 20 trials per task. Increasing data also increases optimization budget, so this is not a fixed-compute comparison.

At $N=100$, SONIC yields atomic/compositional SR of 54/18\% for GR00T N1.7 and 78/29\% for $\pi_{0.5}$. Relative to $N=50$, only $\pi_{0.5}$ atomic performance improves. The extension therefore provides no consistent gain across policies and task categories and is excluded from the five-controller averages.

For $\pi_{0.5}$, action dimensions with quantile-normalized values beyond $|z|=5$ use min/max normalization instead. Controller, scale, and source evaluations use seeds 0--19. A task's shared horizon is the maximum demonstration-derived limit across its data sources.

\subsection{Data-Source Mixtures}
The source study trains GR00T N1.7 and $\pi_{0.5}$ jointly on 16 atomic and four compositional tasks in six scene instances, then tests the same task identities and scenes. The six conditions use teleoperation (T), motion synthesis (S), augmentation (A), and their listed mixtures. Mixtures concatenate selected sources without count matching or source-balanced sampling; GR00T N1.7 samples valid action-window starts and $\pi_{0.5}$ samples frame starts.

\subsection{Motion-Generator Adaptation}
\label{app:generation}
Kimodo generates three-second, 30~Hz G1 motion references and frozen SONIC executes them at 50~Hz in MuJoCo with a 2~ms physics step. Training batches contain 64 condition groups with four candidate motions each: 32 random spatial constraints and 32 task-derived stages. Six denoising steps include four stochastic transitions used for a clipped, group-normalized policy-gradient update~\cite{black2024ddpo,schulman2017ppo}; the simulator and controller are not differentiated through. Gaussian transition log probabilities are normalized by the number of entries in the motion sequence before their differences are exponentiated to form PPO ratios. Full fine-tuning updates the denoiser at $10^{-7}$, while LoRA~\cite{hu2022lora} uses rank 16, scale 32, and a $10^{-5}$ learning rate. The evaluated Full and LoRA checkpoints are from iterations 3,075 and 1,225, respectively. Both use seed 0, frozen SONIC and text encoder, and a trained value head. Validation uses 64 fixed conditions every 25 iterations (seed 9999). For each run, selection uses the highest validation reward among saved checkpoints, independently of the expanded test.

\begin{figure*}[!htbp]
  \centering
  \begin{minipage}[c]{0.56\textwidth}
    \centering
    \includegraphics[width=\linewidth]{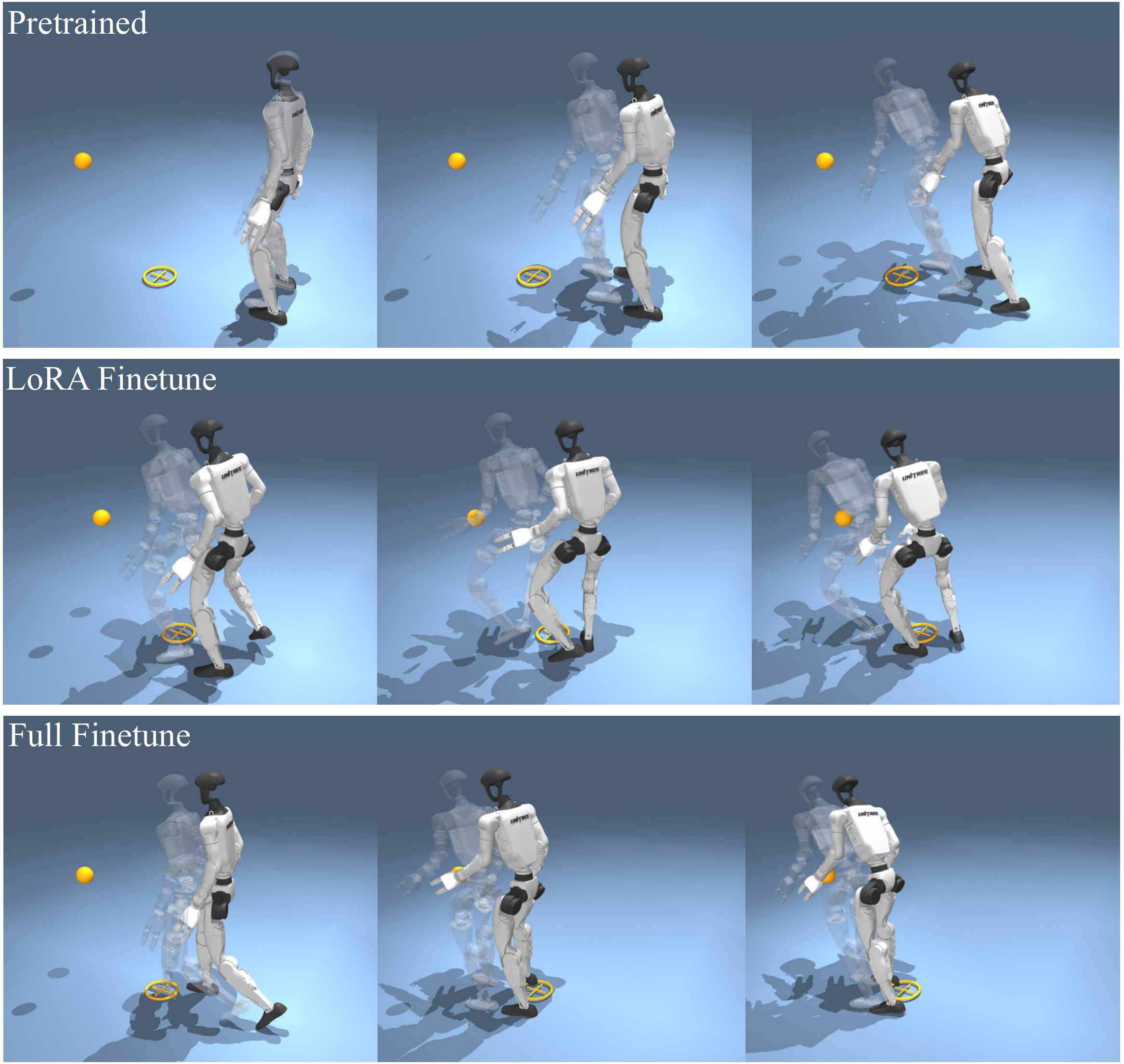}
    \caption{Kimodo locomotion-and-reaching comparison under SONIC execution: walking forward while reaching a shared target with the left hand.}
    \label{fig:kimodo_ft}
  \end{minipage}\hfill
  \begin{minipage}[c]{0.40\textwidth}
    \centering
    \includegraphics[width=\linewidth]{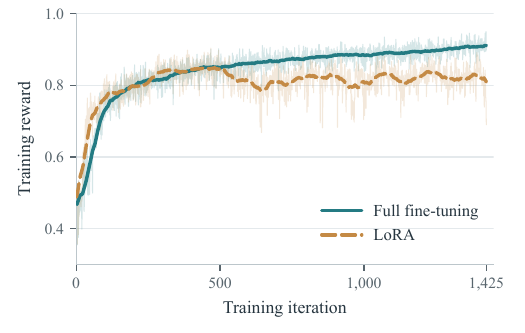}
    \caption{Kimodo training reward over the first 1,425 iterations. Faint curves show per-iteration rewards, and heavier curves show 51-iteration centered moving averages.}
    \label{fig:kimodo_training_reward}
  \end{minipage}
\end{figure*}

Fig.~\ref{fig:kimodo_training_reward} compares training rewards over the first 1,425 iterations of each run. LoRA improves faster initially, but its later reward fluctuates around a plateau, whereas full fine-tuning continues to improve. Over the last 100 shown iterations, mean reward is 0.908 for Full and 0.818 for LoRA. This trend is consistent with the higher overall task success of the selected Full model in Table~\ref{tab:kimodo_ft}.

\label{app:synthesis_provenance}
The synthesis corpus used in the source-mixture study is generated with fully fine-tuned Kimodo models and a frozen SONIC controller that executes the generated motions. Only rollouts that successfully complete the task are retained as training demonstrations.

The augmentation source uses one teleoperated trajectory per task to generate local variations in execution timing and initial conditions. Body motion tokens and finger targets are resampled together, varying total duration by up to 4\% and changing local playback speed while preserving motion order and hand--body synchronization. Robot and movable-object positions vary within the native reset ranges of 5~cm and 2~cm. Each candidate is executed through SONIC and screened for task success and physical validity. The 1,000 accepted episodes came from 1,606 attempts.

\subsection{Kimodo Evaluation Conditions}
\label{app:kimodo_tasks}
The reported scene-task evaluation covers 30 task--scene pairs, comprising ten Approach, five compositional Approach, and fifteen Close tasks, with ten episodes per pair and model. Overall success gives equal weight to these 30 pairs. Thirteen pairs overlap the task-derived fine-tuning set.

The separate goal-reaching evaluation has 198 fixed conditions per model: 54 locomotion, 72 stationary manipulation, and 72 combined locomotion/reaching. Each motion executes for three seconds; root error is planar distance to the shared root target and palm error is 3-D distance between the executed grasp center and shared palm target. Means and sample standard deviations summarize conditions, not independent training runs. Prompt and target ranges overlap training support, so this evaluation is in-distribution.

\section{Simulation and Hardware Evaluation}
\label{app:hardware}
The real-to-sim study covers fridge opening, cup transport from a shelf to a desk, and moving three conveyor boxes to a desk. Simulation uses 50 teleoperation episodes per task. GR00T N1.7, Fast-WAM, and XR-1 are each trained separately per task for 10k updates with seed 42 and budget-end checkpoint selection. GR00T N1.7 uses batch 64 and Fast-WAM batch 32, both with cosine decay from $10^{-4}$. XR-1 retains its main-benchmark recipe, with batch 32 and cosine decay from $2\times10^{-5}$. This fixed budget is distinct from the main experiment's two-pass planning rule.

Evaluation uses one $320\times240$ head camera, SONIC at 50~Hz, seeds 0--19, and no robot/object reset randomization. Limits are 592/951/1,889 control steps, without fall-based early termination. All models start from identical initial states for each seed. Simulation and hardware scenes are approximate counterparts rather than geometrically paired.

\begin{figure}[htbp]
  \centering
  \begin{minipage}[b]{0.46\columnwidth}
    \centering
    \includegraphics[height=3.8cm]{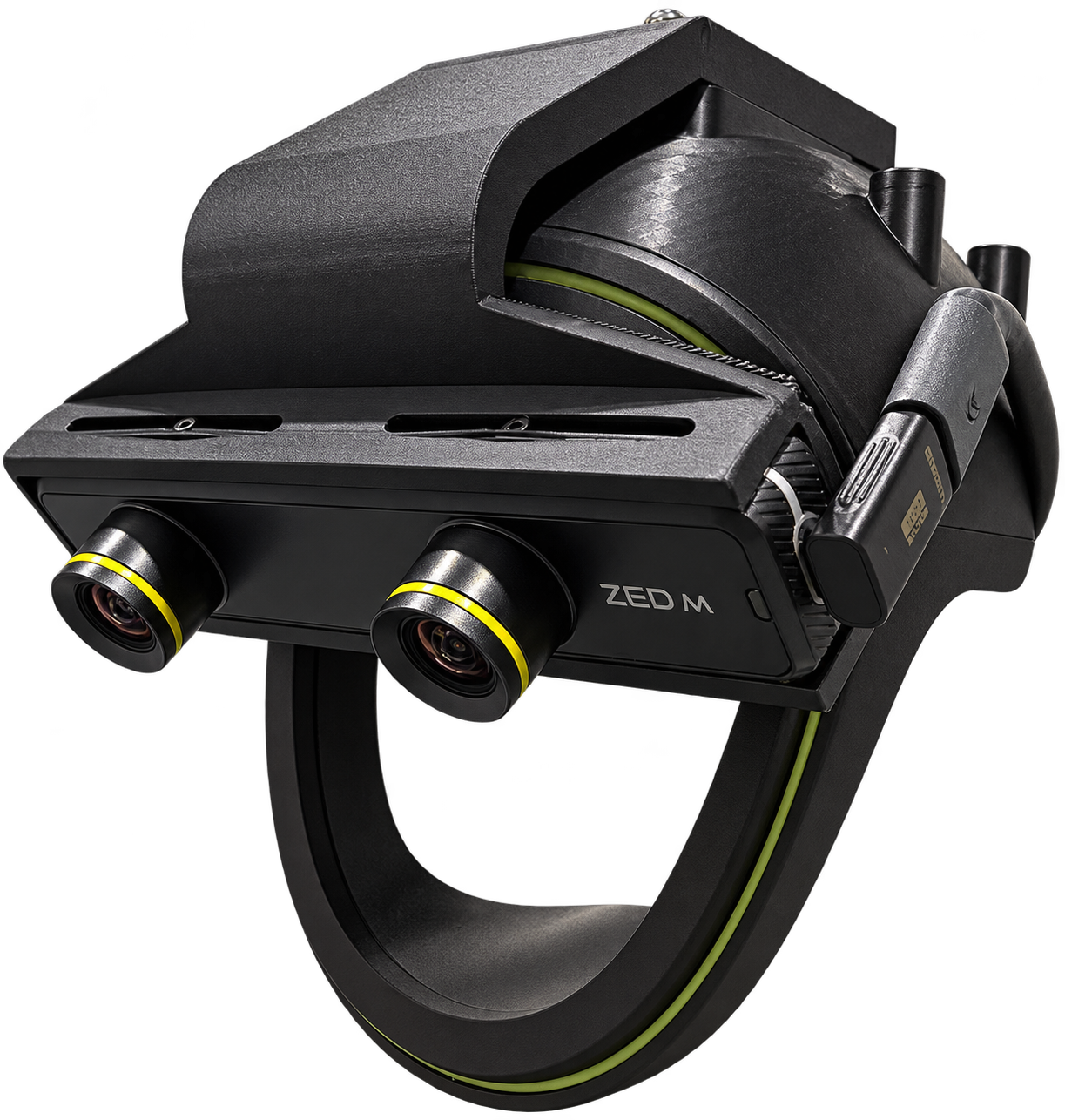}\\[2pt]
    {\footnotesize (a) Front view}
  \end{minipage}\hfill
  \begin{minipage}[b]{0.52\columnwidth}
    \centering
    \includegraphics[height=3.8cm]{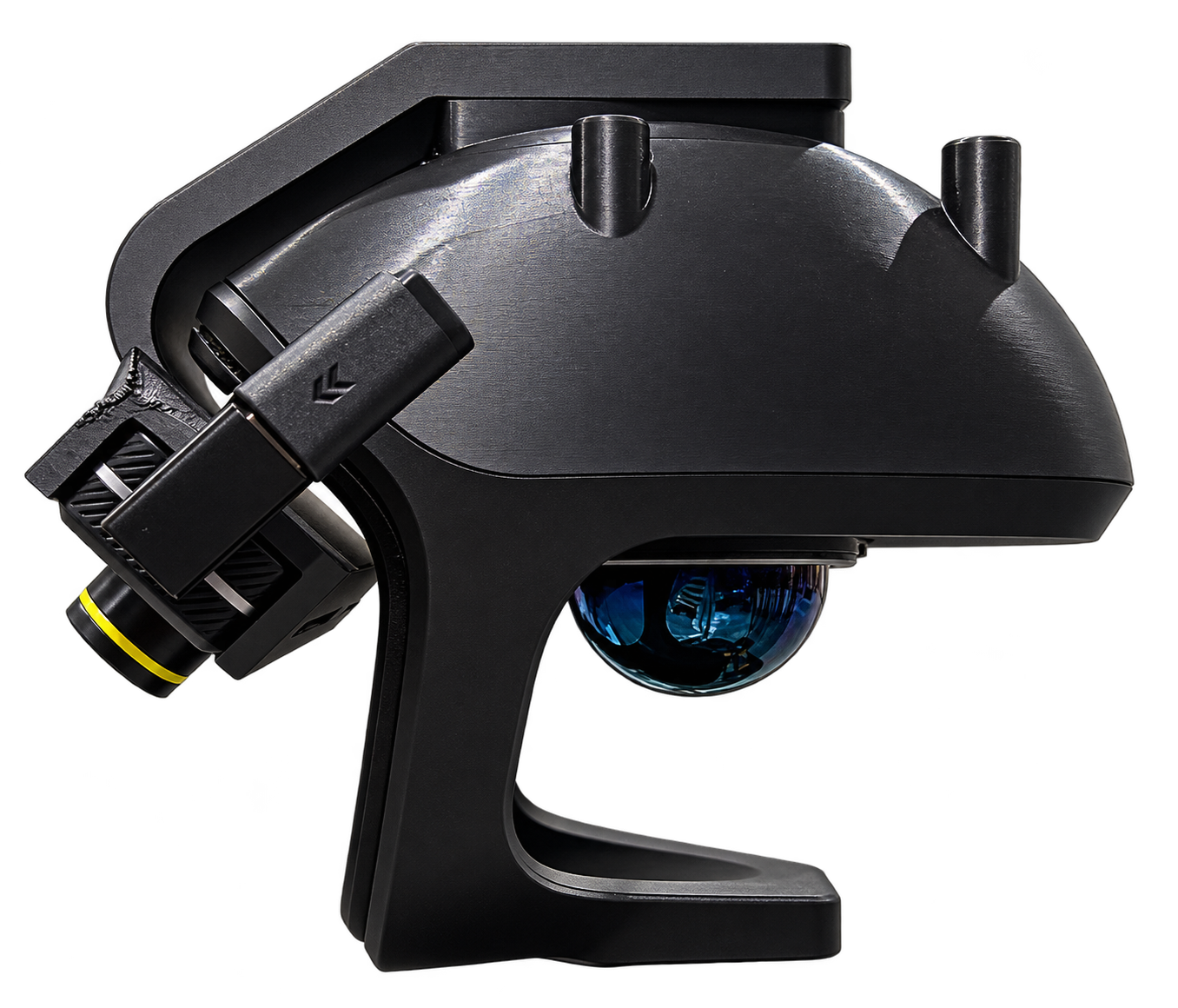}\\[2pt]
    {\footnotesize (b) Side view}
  \end{minipage}
  \caption{Head camera used in the real-robot experiments. (a) Front and (b) side views of the ZED Mini stereo camera mounted on the G1 head; policies use its left RGB image.}
  \label{fig:tro_head_camera}
\end{figure}

In simulation, policies use only the G1  built-in head camera, without wrist cameras. On hardware, policies use the left RGB image of a ZED Mini stereo camera mounted on the head, as shown in Fig.~\ref{fig:tro_head_camera}.

The real-robot teleoperation dataset contains 567 episodes totaling approximately 9.0 hours across fridge opening, cup transport, bottle grasping, and conveyor handling. Hardware training uses 50 demonstrations per evaluated task. Each model is trained separately for each task and follows the training and inference configuration of its corresponding simulation experiment, with real demonstrations replacing simulated data.

Hardware uses a Unitree G1 and human-judged completion. The cup must be placed on the table, the fridge door opened, or a box picked up from the conveyor. Each model--task cell has ten trials. In Fridge/Cup/Conveyor order, success counts are GR00T N1.7 5/6/5, Fast-WAM 4/5/5, and XR-1 8/7/6. Simulation's conveyor criterion requires all three boxes on the desk, whereas the hardware criterion is a box picked from the moving belt. These outcomes therefore use different success predicates.

\end{document}